\documentclass{article} %
\usepackage{colm2024_conference}
\usepackage{graphicx}
\usepackage{microtype}
\usepackage{hyperref}
\usepackage{url}
\usepackage{booktabs}
\usepackage{xcolor}
\usepackage{marvosym}
\usepackage{color,soul}
\usepackage{tabularx}
\usepackage[utf8]{inputenc} %
\usepackage[T1]{fontenc}    %
\usepackage{hyperref}       %
\usepackage{url}            %
\usepackage{booktabs}       %
\usepackage{amsfonts}       %
\usepackage{nicefrac}       %

\usepackage{microtype}      %
\usepackage{multirow}
\usepackage{booktabs, multirow, array, makecell, caption}
\usepackage{multicol}
\definecolor{beaublue}{rgb}{0.74, 0.83, 0.9}
\newcommand{\hlc}[2][yellow]{{%
    \colorlet{foo}{#1}%
    \sethlcolor{foo}\hl{#2}}%
}

\title{Chain-of-Symbol Prompting for Spatial Reasoning in Large Language Models}

\author{%
\textbf{Hanxu Hu}$^{1, 3}\thanks{Equal Contribution}$, 
 \textbf{Hongyuan Lu}$^{2*}\thanks{Corresponding Author}$, \textbf{Huajian Zhang}$^3$, \textbf{Yun-Ze Song}$^1$, \textbf{Wai Lam}$^2$, \textbf{Yue Zhang}$^1$ \\
$^1$Westlake University \quad $^2$The Chinese University of Hong Kong \quad $^3$University of Edinburgh\\
\texttt{huhanxu1233@gmail.com}\\
\texttt{\{hylu,wlam\}@se.cuhk.edu.hk}\\
\texttt{v1hzha17@exseed.ed.ac.uk} \\
\texttt{zhangyue@westlake.edu.cn}
}

\colmfinalcopy %
\begin{document}

\maketitle

\begin{abstract}
In this paper, we first investigate the performance of LLMs on complex planning tasks that require LLMs to understand a virtual spatial environment simulated via natural language and act or reason correspondingly in text. By evaluating on classic spatial planning scenarios, we found that current  LLMs still lack abilities to handle spatial relationships in texts. This arises a question: Is the natural language the best way to represent complex spatial environments for LLMs, or are other alternatives such as symbolic representations more efficient and effective for LLMs? To this end, we propose a novel method called \textsc{CoS} (\textbf{C}hain-\textbf{o}f-\textbf{S}ymbol Prompting) that represents the spatial relationships with condensed symbols during the chained intermediate thinking steps. \textsc{CoS} is easy to use and does not need additional training on LLMs. Extensive experiments indicate that \textsc{CoS} clearly surpasses the performance of the Chain-of-Thought (CoT) Prompting described in natural language in all three spatial reasoning and planning tasks with even fewer tokens used in the inputs compared with CoT. The performance gain is strong, by up to 60.8\% accuracy (from 31.8\% to 92.6\%) on Brick World for GPT-3.5-Turbo. \textsc{CoS} also reduces the number of tokens in the prompt obviously, by up to 65.8\% of the tokens (from 407 to 139) for the intermediate steps from demonstrations on Brick World. Interestingly, we also observed \textbf{emergent ability} of abstract symbols understanding when the size of models scales up. \footnote{Our code are available at \url{https://github.com/hanxuhu/chain-of-symbol-planning}. }
\end{abstract}

\begin{figure*}[h!]
\begin{center}
\vspace{0mm}
\centerline{
\includegraphics[width=14cm]{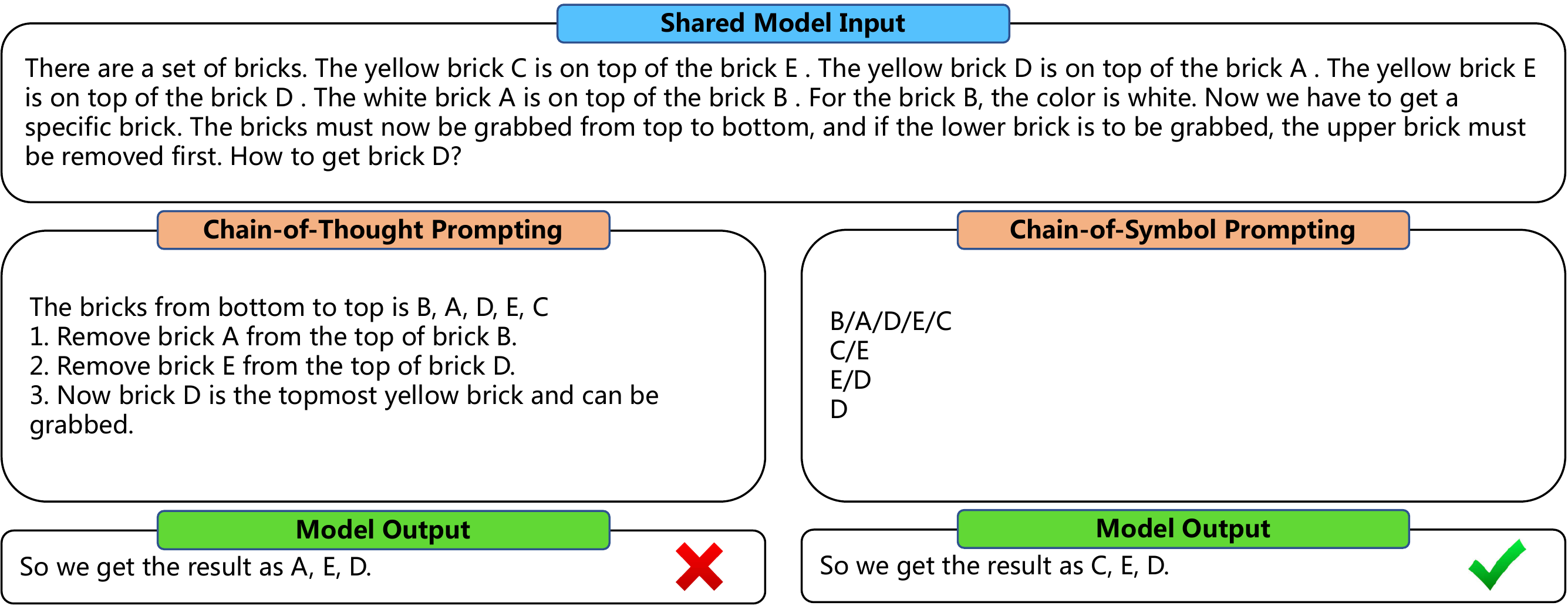}}
\caption{An example for comparison between Chain-of-Thought (CoT) and Chain-of-Symbol (\textsc{CoS}) that elicits large language models in tackling complex planning tasks with higher performance and fewer input tokens. We let the model generate CoT/\textsc{CoS} during inference in a few-shot manner. Results were taken in May 2023 with ChatGPT and can be subject to change.}
\label{cos1}
\end{center}
\vspace{-5mm}
\end{figure*}

\section{Introduction}
Given a set of target behaviour examples, large language models (LLMs) demonstrate exceptional abilities to accomplish a wide range of tasks, frequently exhibiting performance that surpasses that of humans \citep{NEURIPS2020_1457c0d6,2022arXiv220604615S}. 
Specifically, LLMs exhibit impressive sequential textual reasoning ability during inference, resulting in a significant boost in their performance when encountered with reasoning questions described in natural languages \citep{2021arXiv211200114N,wei2022chain}. This phenomenon can be clearly observed with a multi-step chain of intermediate thinking procedure, i.e., a "Chain of Thought" (CoT, \citealt{wei2022chain}).    
\par
Conventional CoT usually leverages \textbf{natural languages} as intermediate thinking steps in prompting. Although CoT can enhance LLMs' ability in many cases, redundant natural languages and irrelevant information also can hamper the performance of LLMs \citep{2023arXiv230200093S} in some cases. For example, spatial languages and descriptions can be hard for language models to understand \citealt{mirzaee-etal-2021-spartqa, SQA} due to complex spatial relationships. Aligning symbols and representing spatial relationships by symbols in word sequences can be a neater representation and thus can be potentially easier to understand by LLMs. We thus explore the use of \textbf{symbols} for LLM prompting, which is still an understudied topic. This is important to study which implies understanding abilities beyond language models for language understanding per se. 
\par 
To explore the role of symbolic representations in prompting, we take the complex spatial understanding and planning as the evaluation scenarios, which require LLMs to understand the virtual spatial environments described through natural language as well as planning and achieving certain goals in such environments. Inspired by existing classic planning competitions and spatial reasoning datasets, we present three domains: (i) Brick World (ii) NLVR-based Manipulation and (iii) Natural Language Navigation. Figure \ref{cos1} illustrates an example for Brick World 1D, and all these three tasks are described in detail in Section \ref{nlp}. These three tasks are all described in natural language. And we also evaluate one existing spatial question answering dataset SPARTUN \citep{SQA} which uses human-generated questions thus closer to realistic situations. For these tasks, LLMs need to understand a virtual environment in natural language, with the spatial relationship between the objects to be operated on and the restrictions on the operation, which is easy for real humans. However, we found that there are still places for improvement in the performance of LLMs on the tasks. 
\par
As a major contribution to this study, we investigate the symbolic representations for spatial relationships, and propose a novel method called \textbf{C}hain-\textbf{o}f-\textbf{S}ymbol (\textsc{CoS}) prompting to elicit spatial understanding and planning abilities on LLMs. As in Figure \ref{cos1}, instead of using intermediate thinking steps described in \textbf{natural language} in CoT prompts shown on the left-hand side, the CoS prompts remove the redundant text description but only using a set of \textbf{symbols} to represent spatial relationships between objects in complex environments. \textsc{CoS} achieves noticeable improvement in both \textbf{performance} and \textbf{efficiency} (by up to 60.8\% improvements in accuracy and 65.8\% for the number of input tokens). We speculate that such an improvement is benefited by the more efficient symbolic representation produced by \textsc{CoS}.
Our main contributions are three-fold:
\begin{itemize}
\setlength\itemsep{0em}
\item 

We evaluate LLMs on both existing classic spatial understanding tasks and our proposed synthetic spatial planning tasks. We spot that there is still room for performance improvements on current LLMs even with CoT.
\item We propose a novel method called \textsc{CoS}, which prompts LLMs to convert the complex environment described with natural language into condensed symbolic representations. 
\textsc{CoS} drastically improves LLMs on the spatial tasks. The accuracy gain of \textsc{CoS} is large, also with a good reduction in the token consumption for LLMs.
\item We conduct an in-depth analysis on \textsc{CoS} to explore the effect of using different symbols, on different LLMs, and different languages to show the robustness of our method.
\end{itemize}

\begin{figure*}[t!]
\begin{center}
\vspace{0mm}
\centerline{
\includegraphics[width=12cm]{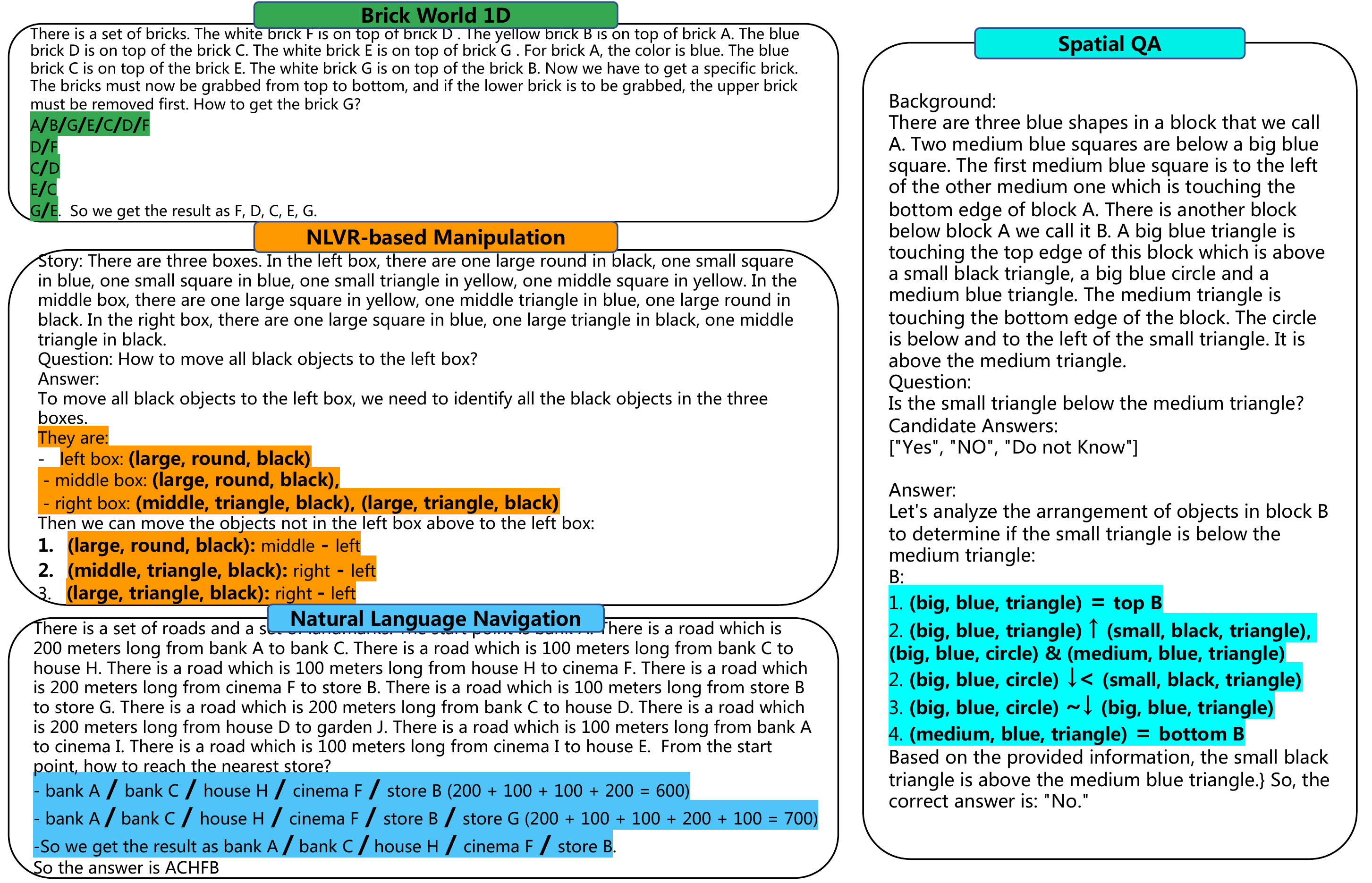}}
\caption{<input, Chain of Symbol, output> 
 example triples for our three proposed tasks: Brick World, NLVR-based Manipulation, and Natural Language Navigation, and SPARTUN dataset \citep{SQA}. Chains of Symbols are highlighted.}
\label{cos2}
\end{center}
\vspace{-5mm}
\end{figure*}

\section{Spatial Planning and Understanding tasks}

\subsection{Natural Language Spatial Planning} 
\label{nlp}
Inspired by classic planning domains and tasks described in \cite{llm+p} and existing spatial reasoning dataset \cite{nlvr}, we explore the performance of LLMs in three natural language spatial planning tasks. For all three tasks, we can formulate the problem as given a virtual scenario described by natural language, and a planning question. LLMs should take both the scenario and the question as the input and output correspondingly to solve the question. Such a solution usually contains a series of steps to achieve a final goal. The final test tasks consist of 5,500 evaluation instances, with 4,000 from Brick World, 1,000 from NLVR-based Manipulation, and the remaining 500 from Natural Language Navigation. We use code to generate these instances based on definition of each task.

\subsection{Brick World}
\label{brickworlds}
Figure \ref{cos2} demonstrates an instance for Brick World (top), which requires the LLMs to acquire certain bricks by grabbing the bricks sequentially. We explore 1D and 2D scenarios for the Brick Worlds task. Specifically, in the 1D scenario, the relationship between bricks is only vertical. In the 2D scenario, in addition to the vertical relationship, there is also a horizontal relationship, which we express as "in the front of". To explore the characteristics of language understanding from LLMs, we investigate different levels of difficulty in the way of describing virtual scenarios. We describe them in increasing levels of difficulty as below. 
\begin{itemize}
\setlength\itemsep{0em}
\item Firstly, we explore labelling bricks from A to Z according to the order of spatial stacking from bottom to top, and the corresponding texts are also described in order from bottom to top, we call this setting "No shuffle". 
\item Secondly, we shuffle the order of the corresponding natural language description while maintaining the labelling rules in alphabetic order called "Shuffle description". 
\item Thirdly, we shuffled the order of labelling so that the spatial relationships do not correspond to the alphabetic order anymore, but are still described in the order from bottom to top in the text description, called "Shuffle label". 
\item Finally, we shuffled both the order of labelling and description. We call it "Shuffle both". 
\end{itemize}
We use colors to represent the bricks, which enriches the information and increases the difficulty of the tasks. For each setting with 1D and 2D, we create 500 evaluation instances. The final evaluation set consists of 4,000 instances.
\subsection{NLVR-based Manipulation}

Figure \ref{cos2} demonstrates an instance for NLVR-based Manipulation (middle). We convert the format of Natural Language Visual Reasoning (NLVR, \cite{nlvr}) tasks into a text-based planning task. Based on the creation rules of synthetic images of NLVR, we create 1,000 natural language descriptions for the virtual spatial environments using Python code. Specifically, for each description, we set three boxes just like NLVR, in the left, middle, and right, and in each box, and there are several objects. Each object has three properties: color, shape, and size. Each description has one related question, the question is about how to move all objects that satisfy a certain condition of one property (such as "all objects in black" or "all rounds") to a specific target box. The ground truth is the set of all objects satisfied with this condition which needs to be moved (not in the target boxes).
\subsection{Natural Language Navigation}
Figure \ref{cos2} demonstrates an instance for Natural Language Navigation (bottom).
Inspired by Vision-and-Language navigation \citep{gu-etal-2022-vision}, we create a virtual spatial environment that is similar to a 2D map of navigation tasks but using natural language description only. Specifically, we define a set of landmarks: $'store', 'bank', 'house', 'cinema', 'garden', 'school'$. For each description, there are 7 to 10 landmarks. We create 500 evaluation instances using Python code: the relationship between landmarks is a binary tree structure, with a root node which indicates the start point in the virtual scenario, and each node other than the leaf nodes has one or two child nodes, with a distance of 100 meters or 200 meters between them. Each description has one related question which is about how to reach the nearest one specific kind of landmark from the starting point.
\subsection{Spatial QA}

We also evaluate \textsc{CoS} on manually annotated existing spatial question answering task,  SPARTUN \citep{SQA}, which contains a larger variety of spatial relation types and spatial expressions compared with previous Spatial QA datasets and our three synthetic spatial planning tasks. And the questions in this dataset are manually annotated, which is closer to real-world scenes. The scenarios in this dataset are described in natural languages based on NLVR \citep{nlvr} and SPARTQA \citep{mirzaee-etal-2021-spartqa}.

\section{Chain-of-Symbol Prompting}
\label{cos-prompt}
We propose Chain-of-Symbol (\textsc{CoS}) prompting for LLMs, which converts the simulated environment with natural language into a condensed symbolic representation that considers spatial relationship. In order to make our constructing method of \textsc{CoS}  generalizable and reliable, we adopt a three-step procedure in creating the demonstrations of our \textsc{CoS} which can be used in any related tasks:
\begin{itemize}
    \item (i) Automatically prompt the LLMs to generate a CoT demonstration in a zero-shot manner
    \item (ii) Correct the generated CoT demonstration if there existing errors.
    \item (iii) Replace the spatial relationships described in natural languages in CoT with random \textbf{symbols}, and only keep objects and symbols, remove other descriptions.
\end{itemize}   We then use the \textsc{CoS} demonstrations to guide the language model in a few-shot manner for prompting LLMs just like CoT \citep{wei2022chain}.
\par 
Figure \ref{cos1} depicts an example of a demonstration of CoS produced by models. In this example, we see that both CoT and \textsc{CoS} receive the same shared simulated spatial environment in natural language texts. \textsc{CoS} depicts a different intermediate thinking process than CoT. The latter represents the environments in a natural language only, while the former use a condensed symbolic representation that considers spatial relationship. Specifically, we use the symbol "\textbf{/}" to represent the spatial relationship "from the top of" here. By doing such a conversion, and removing redundant descriptions, \textsc{CoS} effectively improves the model performance as well as reduces the inference costs with LLMs. 

Figure \ref{cos2} depicts examples of CoS demonstration for all three planning tasks we proposed. For NLVR-based Manipulation, we convert natural language descriptions for objects to the format of a triplet such as "(large, round, black)". For Natural Language Navigation, we represent the order of landmarks by using symbol "/" to connect them. For Spatial QA task, we use a set of symbols such as "=", "$\sim$" to represent different spatial relationships, and use triplet with "( , , )" to represent objects and their attributes.
\par 
CoS prompting has multiple properties that are attractive as a prompting approach for LLMs:
\begin{itemize}
    \item First, \textsc{CoS} effectively allows a neater, shorter, and condensed intermediate procedure than CoT. It is more structured than natural languages, hence easier for human annotators to analyze, check and correct the intermediate thinking process for LLMs.
    \item Second, \textsc{CoS} improves important planning tasks that current LLMs do not tackle well. It provides a better representing method for spatial environments which is easier for LLMs to learn compared with natural language.
    \item Finally, \textsc{CoS} clearly reduces the amount of text input into the LLMs and output from LLMs. This makes it much cheaper to access LLMs with API/GPU.
\end{itemize}

\section{Experiments}
In this section, we first introduce our experimental setup in Section \ref{sec:expset} about the settings of different methods we use, the language models, and the evaluation metrics. Then, in Section \ref{sec:spatialplan}, we report the results of the three spatial planning tasks we proposed. In Section \ref{sec:spatialqa}, we report the results on the SPARTUN dataset. 
\label{s1}
\subsection{Experimental Setup}
\label{sec:expset}
We evaluate CoS and CoT on our proposed three spatial planning tasks and the existing SQA dataset, based on representative LLMs like ChatGPT(gpt-3.5-turbo) and LLAMA-2 series. There are three prompts: zero-shot CoT, few-shot CoT, and few-shot CoS (Ours). 

\paragraph{Zero-shot Chain-of-Thought Prompting} We consider zero-shot CoT as our baseline. The reason is that we have found that our choices of LLMs naturally give their intermediate steps (CoT) in their answers, even without specifically asking them to do so. We also found that asking them to remove the thinking steps obviously degrades the results. Therefore, we allow the LLMs to generate CoT, while we do not put any demonstration to the prompt but give prompts like "Let's think step by step" just as \cite{zsCoT}. For an easier evaluation, we ask the LLMs to output the final results by separating the landmarks with commas.
\paragraph{Chain-of-Thought Prompting} This baseline uses a few-shot CoT, in which we encourage LLMs to think step by step, and we use five demonstrations to guide the LLMs in the thinking procedure. Note that the intermediate thinking procedure is represented as natural language text, just like the Standard Prompting. Like in \citep{wei2022chain}, we manually crafted five demonstrations for each task to guarantee their correctness. To guarantee the consistency and reliability of the prompts, we follow the format of CoT generated by zeroshot-CoT prompting. We use these fixed five demonstrations for evaluations on each task.
\paragraph{Chain-of-Symbol Prompting} As described in Section \ref{cos-prompt}, \textsc{CoS} augments the standard CoT prompting with condensed symbolic representation. While CoT has been shown to give large improvements to LLMs on various tasks \citep{wei2022chain}, we argue that using condensed symbolic representations can be an alternative to describing using natural language texts. We manually converted from CoT demonstrations to CoS using the procedure described in Section \ref{cos-prompt}. Five CoS demonstrations of the same examples with CoT are created for each task of Natural Language Planning.
\paragraph{Language Models} We use both Opensource LLMs (Llama-2) and Closed Source LLMs ChatGPT(Gpt-3.5-turbo) for the evaluation of all tasks. We set the temperature to 0 for all the experiments throughout this paper.
\paragraph{Evaluation Metrics}  For planning tasks, we use three evaluation metrics, namely accuracy, precision, and recall. We define accuracy as the success rate in achieving the final goal. We then compute the Longest Common Sequence (LCS) between the ground truth and LLM output sequence to measure their similarity. We compute precision as the ratio of LCS against the length of the LLM output, and we compute recall as the ratio of LCS against the length of the ground truth. For spatial QA task, we only compute accuracy.

\subsection{Results of Spatial Planning Tasks}
\label{sec:spatialplan}

\subsubsection{Brick World}
\begin{table}[!t]
\vspace{3mm}
\captionof{table}{\label{gpt3.5-bw}
The results of ChatGPT(gpt-3.5-turbo) on Brick World. We report the results with four settings as described in Section \ref{brickworlds}, under both 1D and 2D scenarios. We adopt $N_s=5$, where $N_s$ represents the number of demonstrations for \textsc{CoS} and CoT. The best results are bolded. For \textsc{CoS} and CoT, we report the average and the standard deviation from three runs with different sets of demonstrations. \textbf{Acc.} represents accuracy, \textbf{Pre.} represents precision, and \textbf{Rec.} represents recall. zs-CoT represents zero-shot CoT. We report the average number of tokens in the intermediate steps.
}
\vspace{3mm}
\scriptsize
\setlength\tabcolsep{1.5pt}
\setlength\aboverulesep{0pt}\setlength\belowrulesep{0pt}
\setcellgapes{3pt}\makegapedcells
\begin{tabular*}{\linewidth}{lccccccccccccc}
\hline
\textbf{Model} & \multicolumn{3}{c}{\textbf{No Shuffle}} & \multicolumn{3}{c}{\textbf{Shuffle Description}} & \multicolumn{3}{c}{\textbf{Shuffle Label}} & \multicolumn{4}{c}{\textbf{Shuffle Both}} \\
\hline
& \textbf{Acc.} & \textbf{Pre.} & \textbf{Rec.} &  \textbf{Acc.} & \textbf{Pre.} & \textbf{Rec.} &  \textbf{Acc.} & \textbf{Pre.} & \textbf{Rec.}  &  \textbf{Acc.} & \textbf{Pre.} & \textbf{Rec.} & \textbf{Tok.}\\
\hline
\multicolumn{13}{c}{\textit{1D Scenario}}
\\
\hline
zs-CoT & 61.0&77.2&71.9&60.4&77.5&77.5&31.8&63.4&59.8&28.2&58.6&55.3&-
 \\
CoT &81.0\tiny{$\pm $11.0}&87.7$\pm $4.5&90.1\tiny{$\pm$2.6}&71.5\tiny{$\pm$9.2}&90.7\tiny{$\pm$ 3.6}&81.8\tiny{$\pm$7.1}&75.1\tiny{$\pm$10.1}&88.0\tiny{$\pm$ 3.6}&90.1\tiny{$\pm$0.9}&43.0\tiny{$\pm$4.4}&71.4\tiny{$\pm$3.3}&75.7\tiny{$\pm$1.6}&407
\\
\textsc{CoS} & \textbf{96.6\tiny{$\pm $1.9}}&\textbf{98.0\tiny{$\pm$0.8}}&\textbf{97.7\tiny{$\pm$0.8}}&\textbf{95.9\tiny{$\pm$1.2}}&\textbf{97.9\tiny{$\pm$ 0.6}}&\textbf{97.5\tiny{$\pm$0.3}}&\textbf{92.6\tiny{$\pm$2.0}}&\textbf{97.0\tiny{$\pm$ 1.3}}&\textbf{95.9\tiny{$\pm$1.1}}&\textbf{69.7\tiny{$\pm$5.1}}&\textbf{86.7\tiny{$\pm$4.2}}&\textbf{83.6\tiny{$\pm$1.6}}&\textbf{139}
\\
\hline
\multicolumn{13}{c}{\textit{2D Scenario}}
\\
\hline
zs-CoT & 32.7&53.8&60.6 &
14.8&31.9&46.9 &
13.0&32.0&42.3 &
9.8&30.4&38.4&-
\\
CoT & 25.0\tiny{$\pm$15.6}&49.8\tiny{$\pm$9.8}&45.0\tiny{$\pm$10.5}&
21.5\tiny{$\pm$8.2}&45.6\tiny{$\pm$5.4}&41.2\tiny{$\pm$6.3}&
21.8\tiny{$\pm$2.3}&44.7\tiny{$\pm$5.9}&43.2\tiny{$\pm$4.0}&14.9\tiny{$\pm$3.4}&38.1\tiny{$\pm$2.9}&36.4\tiny{$\pm$3.5}&546\\
\textsc{CoS} & \textbf{60.7\tiny{$\pm$1.9}} &
\textbf{67.2\tiny{$\pm$1.1}} &
\textbf{71.3\tiny{$\pm$1.3}} & 
\textbf{33.7\tiny{$\pm$3.2}} & 
\textbf{46.7\tiny{$\pm$0.8}} &
\textbf{50.0\tiny{$\pm$1.5}} &
\textbf{23.5\tiny{$\pm$5.0}} & 
\textbf{45.9\tiny{$\pm$0.8}} &
\textbf{63.0\tiny{$\pm$12.1}} & 
\textbf{28.9\tiny{$\pm$2.3}} & 
\textbf{46.3\tiny{$\pm$1.0}} &
\textbf{44.4\tiny{$\pm$2.8}} & \textbf{341}
\\
\hline
\end{tabular*}
\vspace{2mm}
\end{table}

Table \ref{gpt3.5-bw} reports the results of \textsc{CoS} against the zs-CoT and CoT on the task of Brick World. First of all, we can see that the complexity increases both from the 1D scenario to the 2D scenario and from the setting of No Shuffle to the setting of Shuffle Both, together with a  drop in the performance. ChatGPT with zs-CoT does not perform well, with only 9.8\% accuracy on the most difficult setting Shuffle Both under the 2D scenario. Although CoT brings some improvements, the performance for CoT is still not satisfying, with an accuracy of 43.0\% which is just below the 50\% bar for setting Shuffle Both under the 1D scenario. In contrast, we see that \textsc{CoS} gives very good improvements on this setting (from 28.2\% to 69.7\%). We found that \textsc{CoS} gives consistent improvements to all the settings on Brick World, clearly surpassing CoT. The largest gain is on the setting of Shuffle Label under the 1D scenario, with 60.8\% improvements in accuracy (from 31.8\% to 92.6\%).

\subsubsection{Further Analysis of Brick World}

\textbf{Randomness in the Experiments} To investigate the randomness in our experiments, we run multiple trials with three different sets of demonstrations for CoT and \textsc{CoS}. Table \ref{gpt3.5-bw} reports their means and standard deviations. We see a general trend here that \textsc{CoS} usually reports a lower standard deviation than CoT (for example, a standard deviation of 1.9 for Acc. for No Shuffle under the 1D scenario for \textsc{CoS}, against 11.0 for CoT). This represents that \textsc{CoS} is more stable than CoT on Brick World.

\begin{minipage}[h!]{0.4\textwidth}

\includegraphics[width=0.87\linewidth]
{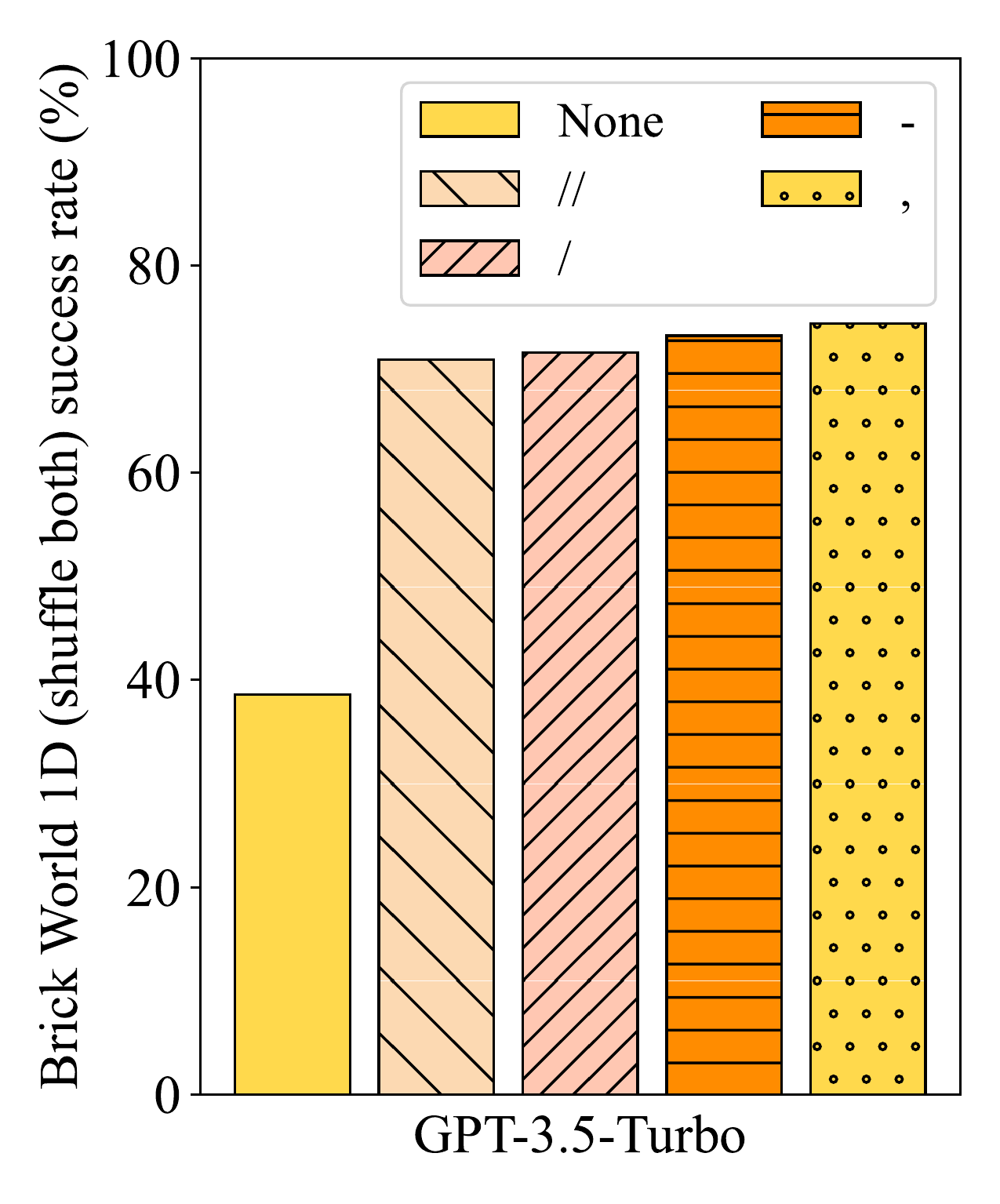}
\captionsetup{width=0.8\linewidth}
\captionof{figure}{\label{fig:ab_sym}Performance of using different symbols for \textsc{CoS} on Brick World 1D (Shuffle Both) in accuracy.}
\vspace{5mm}
\end{minipage} 
\begin{minipage}[h!]{0.6\textwidth}
\textbf{CoS on the Different Language} In addition to the tasks described in English, we also tested \textsc{CoS} on Brick World in Chinese.
CoT reports 22.9\% accuracy, and \textsc{CoS} reports 39.1\% in 1D scenario of Brick World, which demonstrates the robustness of \textsc{CoS} to a language other than English. 
\par
\textbf{Robustness to Different Symbols} Figure \ref{fig:ab_sym} demonstrates the robustness of using different symbols for \textsc{CoS}. As we can see, using different symbols brings consistent performance, while not having any symbol drastically impacts the results. Among all the symbols, the comma gives the best results. We conclude that \textsc{CoS} is robust to the selection of symbol.
\par 
\textbf{Results on Different Language Models} Figure~\ref{scaling} reports the results on LLAMA-2 under the 1D scenario. The experimental results align with our previous conclusion that \textsc{CoS} outperforms CoT obviously.
\par
\textbf{Saving Tokens for Prompting} One advantage featured by \textsc{CoS} is that it reduces the number of input tokens to be fed into LLMs. For Brick World (1D scenario), \textsc{CoS} reduces the
\end{minipage} number of tokens for the intermediate steps from 407 to 139 (Table \ref{gpt3.5-bw}, the numbers are reported from OpenAI Playground\footnote{https://platform.openai.com/playground}). This subsequently saves the costs of accessing LLMs via API/GPUs.
\par 
\subsubsection{NLVR-based Manipulation and Natural Language Navigation}
\label{s2}
For both of these two tasks, we adopt almost the same experimental settings as the ones for Brick World. The only difference is the evaluation metrics we report, we compute precision and recall based on the set rather than the Longest Common Sequence. 
\\
\par
\begin{table}[t!]
\small
\hspace{0.3cm}
\captionof{table}{\label{gpt3.5-nvlr}
The automatic evaluation results with gpt-3.5-turbo on Natural Language Navigation and NLVR-based Manipulation. We set $N_s=5$, where $N_s$ represents the number of demonstrations for prompting with \textsc{CoS} and CoT. The best results are bolded. We report the average and the standard deviation from three runs with different demonstrations. \textbf{Acc.} represents accuracy, \textbf{Pre.} represents precision, and \textbf{Rec.} represents recall (precision and recall are computed with sets in this case). 
}
    \setlength\tabcolsep{6pt}
\setlength\aboverulesep{0pt}\setlength\belowrulesep{0pt}
\setcellgapes{3pt}\makegapedcells
\begin{tabular*}{\linewidth}{@{\extracolsep{\fill}}lcccccccc}
\hline
\textbf{Task} & \multicolumn{4}{c}{\textbf{Natural Language Navigation}} & \multicolumn{4}{c}{\textbf{NLVR-based Manipulation}}\\
\hline
& \textbf{Acc.} & \textbf{Pre.} & \textbf{Rec.} & \textbf{Tok.} & \textbf{Acc.} & \textbf{Pre.} & \textbf{Rec.} & \textbf{Tok.}\\
\hline
zs-CoT & 52.8 &74.0 &79.6 &- &18.6&26.9&19.7&-
\\
CoT & 53.6\scriptsize{$\pm$2.8} & 76.3\scriptsize{$\pm$1.1} &81.7\scriptsize{$\pm$0.8}&390
 & 64.1\scriptsize{$\pm$3.8} &81.7\scriptsize{$\pm$1.3}&84.5\scriptsize{$\pm$0.7}&653\\
\textsc{CoS} & \textbf{74.9\scriptsize{$\pm$3.4}} & \textbf{87.9\scriptsize{$\pm$1.8}} & \textbf{86.7\scriptsize{$\pm$3.0}} & \textbf{239} & \textbf{68.4\scriptsize{$\pm$2.3}} &\textbf{71.2\scriptsize{$\pm$1.9}
}& \textbf{82.9\scriptsize{$\pm$2.1}}  & \textbf{534}
\\
\hline
\end{tabular*}
\end{table}

\textbf{Main Results} Table \ref{gpt3.5-nvlr} reports the results of NLVR-based Manipulation and Natural Language Navigation with GPT-3.5-turbo. For both of these two tasks, \textsc{CoS} reports a higher performance than \textsc{CoT} and zero-shot CoT prompting on all of the metrics.
\par
\textbf{Saving Tokens for Prompting} One advantage of \textsc{CoS} is that it can reduces the number of input tokens. Table \ref{gpt3.5-nvlr} reported that for NLVR-based Manipulation, \textsc{CoS} reduces the number of tokens for the intermediate steps from 653 to 534, nearly by half of the original intermediate steps (we separate the tokens by space). This subsequently saves the costs of accessing LLMs via API/GPU, which enables easier access to the models.

\begin{table}[t]
\centering
\captionof{table}{\label{spqa}
The automatic evaluation results with GPT-3.5-Turbo and GPT-4 on SPARTUN dataset. We apply CoT with 5 shots, and CoS with 5 shots (Ours) respectively. We report the number of tokens in the intermediate steps from demonstrations as the last column.
}
\begin{tabular}{@{}ccccc@{}}
\toprule
Model  & GPT-3.5-Turbo  & GPT-4         & Tokens \\ \midrule
CoT-5  & 47.1                   & 69.8          &    198    \\
CoS-5  & \textbf{49.4}          & \textbf{72.6} &   \textbf{167}    \\ \bottomrule
\end{tabular}
\end{table}

\subsection{Spatial Question Answering}
\label{sec:spatialqa}
We also explore the effectiveness of \textsc{CoS} in a more real-world scenario, by using existing human annotated spatial QA dataset SPARTUN \citep{SQA}. Specifically, we applied both \textsc{CoS} and CoT on GPT-3.5-Turbo and GPT-4. \textsc{CoS} gains better performance and uses fewer tokens compared with CoT. In table \ref{spqa}, we report the results of performance, and both CoT and CoS have 5 shots. It should be noticed that there are far more types of spatial relationships in SPARTUN dataset than our proposed planning tasks, so the results indicate CoS can gain promising performance even when there are a lot of symbols to represent different spatial relationships.

\subsection{Results of different size LLAMA-2}

\begin{figure*}[t!]
\begin{center}
\vspace{0mm}
\centerline{
\includegraphics[width=10cm]{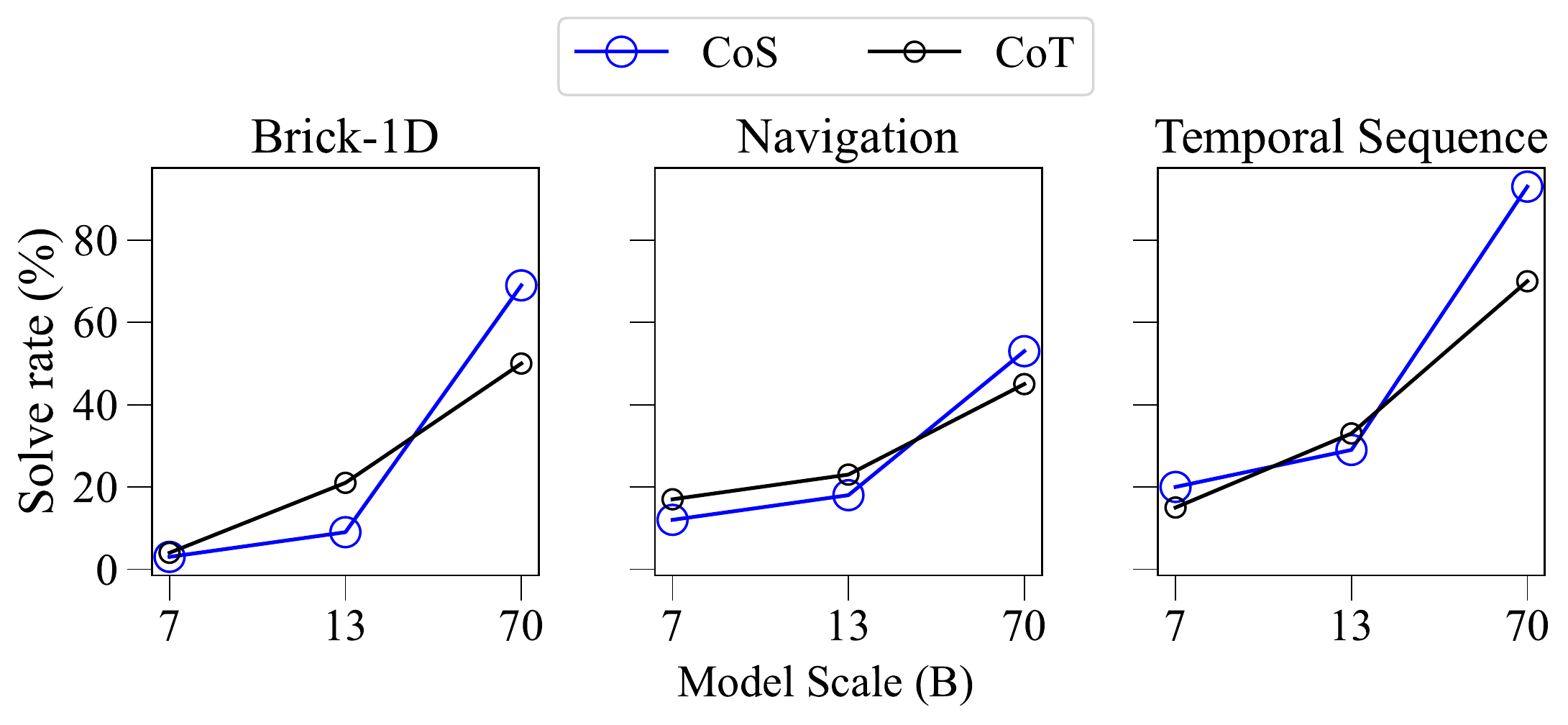}}
\caption{Scaling curve of CoS and CoT of Llama-2 on three tasks.}
\label{scaling}
\end{center}
\vspace{-5mm}
\end{figure*}

We also evaluate CoS on current open-source representative models like LLAMA-2 \cite{touvron2023llama} series to further validate the effectiveness and generality of our method. We use LLAMA-2 with different size (7B, 13B and 13B). As shown in Figure~\ref{scaling}, when the model size is small (7B and 13B), CoS cannot outperform CoT in many cases, but in 70B, CoS gain a clear better performance in all three tasks compared with using CoT. It can be seen that as the model size increases, the ratio of model performance to parameters for CoS has a larger slope compared to CoT. This indicate that large language models might can have emergent ability of abstract symbols understanding.

\section{Related Work}
\paragraph{In-Context Learning} Large language models (LLMs) have demonstrated remarkable few-shot learning abilities across various domains \citep{NEURIPS2020_1457c0d6,2022arXiv220604615S}, which is also called as in-context learning, leading to a paradigm shift in AI to use LLMs as foundational models for language-related tasks, either directly or through fine-tuning \citep{2021arXiv210807258B,hu2022lora}. While less relevant to \textsc{CoS}, a concurrent work converts natural language into executable actions for robots with ChatGPT \citep{wake2023chatgpt}. Another very recent concurrent work uses Symbol Tuning that replaces natural language labels with arbitrary symbols to improve in-context learning \citep{2023arXiv230508298W}.
\paragraph{Chain-of-Thought Reasoning}
The ability of LLMs \citep{NEURIPS2020_1457c0d6,2022arXiv220604615S} to perform complex reasoning tasks can be significantly enhanced by using a show known as Chain-of-Thought (CoT) prompting, which involves providing them with intermediate reasoning steps \citep{2021arXiv211200114N,wei2022chain}. Such a phenomenon also generalizes to the multilingual settings \citep{shi2023language}. Despite the fact that CoT is powerful, there are reports that demonstrate that CoT is not always useful and that integrating CoT degrades the performance on the task of Machine Translation in their experiment. And this is possibly due to the word-by-word translation \citep{2023arXiv230313780P}.
\paragraph{Spatial Reasoning}
Spatial reasoning over natural language texts has been an important research direction in the community \citep{janner-etal-2018-representation, mirzaee-etal-2021-spartqa}. \citet{janner-etal-2018-representation} proposes to leverage representation learning on a navigation task that requires the agent to move a specific location. \citet{rojowiec-etal-2020-generating} proposes a new task on spatial reasoning that requires the language model to generate natural language instructions for `before' and `after' image pairs. \citet{mirzaee-etal-2021-spartqa} proposes a new benchmark for spatial question-answering with `which' and `what' questions regarding the environment. In a concurrent work, \citet{2023arXiv230402868T} demonstrates that LLMs perform poorly on text-based games with question-answering tasks that require several steps of reasoning. 
\paragraph{Navigation and Path Planning}
Language grounding navigation \citep{gu-etal-2022-vision} refers to the interdisciplinary task that requires the intelligent agent to perceive the visual environment and guide the user to the goal location through natural language instructions \citep{8954259,8954308}. Path planning \citep{PANOV2018347,KRISHNALAKSHMANAN2020103078} refers to the tasks that require the agent to plan its own path to achieve certain goals such as the shortest path or maximizing the cleaning area, typically through the use of reinforcement learning. These areas are highly relevant to the spatial planning tasks we explored and \textsc{CoS}, as the spatial environments can be potentially represented by symbolic representations. We leave the investigations of these application areas to future studies. \citep{sun2024survey}

\section{Conclusion}
 We found that current popular LLMs still lack abilities in complex spatial planning and understanding tasks. To this end, we propose a novel method called \textsc{CoS} (\textbf{C}hain-\textbf{o}f-\textbf{S}ymbol Prompting) that converts spatial relationships described in natural languages to condensed symbolic representations in the chained intermediate thinking steps. \textsc{CoS} is easy to use and does not need additional training on LLMs. Extensive experiments indicate that using few-shot \textsc{CoS} demonstration clearly surpasses the performance of using CoT described in natural languages on all three spatial planning tasks we proposed and the representative spatial QA benchmark with even fewer tokens (down to about 1/3 tokens of the thinking steps with CoT) used in the inputs compared with CoT prompting. The performance gain is strong, by up to 60.8\% accuracy (from 31.8\% to 92.6\%) on Brick World for ChatGPT.
\paragraph{Limitations}
Refer to the Appendix for the section on Broader Impact. In addition, we only use two models to verify the effectiveness of our method due to the limited time and resources. It would be interesting to apply our method to more models with different sizes to see whether there is an emergent ability of CoS for LLMs. Nevertheless, our choices of foundation models are representative and they are popular LLMs.

\section*{Acknowledgments}
We acknowledge funding support from the NSFC Key project 62336006, and Center for Perceptual and Interactive Intelligence (CPII) Ltd. under the Innovation and Technology Commission's innoHK scheme.

\bibliography{colm2024_conference}

\begin{thebibliography}{26}
\providecommand{\natexlab}[1]{#1}
\providecommand{\url}[1]{\texttt{#1}}
\expandafter\ifx\csname urlstyle\endcsname\relax
  \providecommand{\doi}[1]{doi: #1}\else
  \providecommand{\doi}{doi: \begingroup \urlstyle{rm}\Url}\fi

\bibitem[{Bommasani} et~al.(2021){Bommasani}, {Hudson}, {Adeli}, {Altman}, {Arora}, {von Arx}, {Bernstein}, {Bohg}, {Bosselut}, {Brunskill}, {Brynjolfsson}, {Buch}, {Card}, {Castellon}, {Chatterji}, {Chen}, {Creel}, {Quincy Davis}, {Demszky}, {Donahue}, {Doumbouya}, {Durmus}, {Ermon}, {Etchemendy}, {Ethayarajh}, {Fei-Fei}, {Finn}, {Gale}, {Gillespie}, {Goel}, {Goodman}, {Grossman}, {Guha}, {Hashimoto}, {Henderson}, {Hewitt}, {Ho}, {Hong}, {Hsu}, {Huang}, {Icard}, {Jain}, {Jurafsky}, {Kalluri}, {Karamcheti}, {Keeling}, {Khani}, {Khattab}, {Koh}, {Krass}, {Krishna}, {Kuditipudi}, {Kumar}, {Ladhak}, {Lee}, {Lee}, {Leskovec}, {Levent}, {Li}, {Li}, {Ma}, {Malik}, {Manning}, {Mirchandani}, {Mitchell}, {Munyikwa}, {Nair}, {Narayan}, {Narayanan}, {Newman}, {Nie}, {Niebles}, {Nilforoshan}, {Nyarko}, {Ogut}, {Orr}, {Papadimitriou}, {Park}, {Piech}, {Portelance}, {Potts}, {Raghunathan}, {Reich}, {Ren}, {Rong}, {Roohani}, {Ruiz}, {Ryan}, {R{\'e}}, {Sadigh}, {Sagawa}, {Santhanam}, {Shih}, {Srinivasan}, {Tamkin}, {Taori},
  {Thomas}, {Tram{\`e}r}, {Wang}, {Wang}, {Wu}, {Wu}, {Wu}, {Xie}, {Yasunaga}, {You}, {Zaharia}, {Zhang}, {Zhang}, {Zhang}, {Zhang}, {Zheng}, {Zhou}, and {Liang}]{2021arXiv210807258B}
Rishi {Bommasani}, Drew~A. {Hudson}, Ehsan {Adeli}, Russ {Altman}, Simran {Arora}, Sydney {von Arx}, Michael~S. {Bernstein}, Jeannette {Bohg}, Antoine {Bosselut}, Emma {Brunskill}, Erik {Brynjolfsson}, Shyamal {Buch}, Dallas {Card}, Rodrigo {Castellon}, Niladri {Chatterji}, Annie {Chen}, Kathleen {Creel}, Jared {Quincy Davis}, Dora {Demszky}, Chris {Donahue}, Moussa {Doumbouya}, Esin {Durmus}, Stefano {Ermon}, John {Etchemendy}, Kawin {Ethayarajh}, Li~{Fei-Fei}, Chelsea {Finn}, Trevor {Gale}, Lauren {Gillespie}, Karan {Goel}, Noah {Goodman}, Shelby {Grossman}, Neel {Guha}, Tatsunori {Hashimoto}, Peter {Henderson}, John {Hewitt}, Daniel~E. {Ho}, Jenny {Hong}, Kyle {Hsu}, Jing {Huang}, Thomas {Icard}, Saahil {Jain}, Dan {Jurafsky}, Pratyusha {Kalluri}, Siddharth {Karamcheti}, Geoff {Keeling}, Fereshte {Khani}, Omar {Khattab}, Pang~Wei {Koh}, Mark {Krass}, Ranjay {Krishna}, Rohith {Kuditipudi}, Ananya {Kumar}, Faisal {Ladhak}, Mina {Lee}, Tony {Lee}, Jure {Leskovec}, Isabelle {Levent}, Xiang~Lisa {Li}, Xuechen
  {Li}, Tengyu {Ma}, Ali {Malik}, Christopher~D. {Manning}, Suvir {Mirchandani}, Eric {Mitchell}, Zanele {Munyikwa}, Suraj {Nair}, Avanika {Narayan}, Deepak {Narayanan}, Ben {Newman}, Allen {Nie}, Juan~Carlos {Niebles}, Hamed {Nilforoshan}, Julian {Nyarko}, Giray {Ogut}, Laurel {Orr}, Isabel {Papadimitriou}, Joon~Sung {Park}, Chris {Piech}, Eva {Portelance}, Christopher {Potts}, Aditi {Raghunathan}, Rob {Reich}, Hongyu {Ren}, Frieda {Rong}, Yusuf {Roohani}, Camilo {Ruiz}, Jack {Ryan}, Christopher {R{\'e}}, Dorsa {Sadigh}, Shiori {Sagawa}, Keshav {Santhanam}, Andy {Shih}, Krishnan {Srinivasan}, Alex {Tamkin}, Rohan {Taori}, Armin~W. {Thomas}, Florian {Tram{\`e}r}, Rose~E. {Wang}, William {Wang}, Bohan {Wu}, Jiajun {Wu}, Yuhuai {Wu}, Sang~Michael {Xie}, Michihiro {Yasunaga}, Jiaxuan {You}, Matei {Zaharia}, Michael {Zhang}, Tianyi {Zhang}, Xikun {Zhang}, Yuhui {Zhang}, Lucia {Zheng}, Kaitlyn {Zhou}, and Percy {Liang}.
\newblock {On the Opportunities and Risks of Foundation Models}.
\newblock \emph{arXiv e-prints}, art. arXiv:2108.07258, August 2021.
\newblock \doi{10.48550/arXiv.2108.07258}.

\bibitem[Brown et~al.(2020)Brown, Mann, Ryder, Subbiah, Kaplan, Dhariwal, Neelakantan, Shyam, Sastry, Askell, Agarwal, Herbert-Voss, Krueger, Henighan, Child, Ramesh, Ziegler, Wu, Winter, Hesse, Chen, Sigler, Litwin, Gray, Chess, Clark, Berner, McCandlish, Radford, Sutskever, and Amodei]{NEURIPS2020_1457c0d6}
Tom Brown, Benjamin Mann, Nick Ryder, Melanie Subbiah, Jared~D Kaplan, Prafulla Dhariwal, Arvind Neelakantan, Pranav Shyam, Girish Sastry, Amanda Askell, Sandhini Agarwal, Ariel Herbert-Voss, Gretchen Krueger, Tom Henighan, Rewon Child, Aditya Ramesh, Daniel Ziegler, Jeffrey Wu, Clemens Winter, Chris Hesse, Mark Chen, Eric Sigler, Mateusz Litwin, Scott Gray, Benjamin Chess, Jack Clark, Christopher Berner, Sam McCandlish, Alec Radford, Ilya Sutskever, and Dario Amodei.
\newblock Language models are few-shot learners.
\newblock In H.~Larochelle, M.~Ranzato, R.~Hadsell, M.F. Balcan, and H.~Lin (eds.), \emph{Advances in Neural Information Processing Systems}, volume~33, pp.\  1877--1901. Curran Associates, Inc., 2020.

\bibitem[Chen et~al.(2019)Chen, Suhr, Misra, Snavely, and Artzi]{8954308}
Howard Chen, Alane Suhr, Dipendra Misra, Noah Snavely, and Yoav Artzi.
\newblock Touchdown: Natural language navigation and spatial reasoning in visual street environments.
\newblock In \emph{2019 IEEE/CVF Conference on Computer Vision and Pattern Recognition (CVPR)}, pp.\  12530--12539, 2019.
\newblock \doi{10.1109/CVPR.2019.01282}.

\bibitem[Gu et~al.(2022)Gu, Stefani, Wu, Thomason, and Wang]{gu-etal-2022-vision}
Jing Gu, Eliana Stefani, Qi~Wu, Jesse Thomason, and Xin Wang.
\newblock Vision-and-language navigation: A survey of tasks, methods, and future directions.
\newblock In \emph{Proceedings of the 60th Annual Meeting of the Association for Computational Linguistics (Volume 1: Long Papers)}, pp.\  7606--7623, Dublin, Ireland, May 2022. Association for Computational Linguistics.
\newblock \doi{10.18653/v1/2022.acl-long.524}.
\newblock URL \url{https://aclanthology.org/2022.acl-long.524}.

\bibitem[Hu et~al.(2022)Hu, yelong shen, Wallis, Allen-Zhu, Li, Wang, Wang, and Chen]{hu2022lora}
Edward~J Hu, yelong shen, Phillip Wallis, Zeyuan Allen-Zhu, Yuanzhi Li, Shean Wang, Lu~Wang, and Weizhu Chen.
\newblock Lo{RA}: Low-rank adaptation of large language models.
\newblock In \emph{International Conference on Learning Representations}, 2022.

\bibitem[Janner et~al.(2018)Janner, Narasimhan, and Barzilay]{janner-etal-2018-representation}
Michael Janner, Karthik Narasimhan, and Regina Barzilay.
\newblock Representation learning for grounded spatial reasoning.
\newblock \emph{Transactions of the Association for Computational Linguistics}, 6:\penalty0 49--61, 2018.
\newblock \doi{10.1162/tacl_a_00004}.
\newblock URL \url{https://aclanthology.org/Q18-1004}.

\bibitem[Kojima et~al.(2022)Kojima, Gu, Reid, Matsuo, and Iwasawa]{zsCoT}
Takeshi Kojima, Shixiang~Shane Gu, Machel Reid, Yutaka Matsuo, and Yusuke Iwasawa.
\newblock Large language models are zero-shot reasoners.
\newblock \emph{Advances in neural information processing systems}, 35:\penalty0 22199--22213, 2022.

\bibitem[{Krishna Lakshmanan} et~al.(2020){Krishna Lakshmanan}, {Elara Mohan}, Ramalingam, {Vu Le}, Veerajagadeshwar, Tiwari, and Ilyas]{KRISHNALAKSHMANAN2020103078}
Anirudh {Krishna Lakshmanan}, Rajesh {Elara Mohan}, Balakrishnan Ramalingam, Anh {Vu Le}, Prabahar Veerajagadeshwar, Kamlesh Tiwari, and Muhammad Ilyas.
\newblock Complete coverage path planning using reinforcement learning for tetromino based cleaning and maintenance robot.
\newblock \emph{Automation in Construction}, 112:\penalty0 103078, 2020.
\newblock ISSN 0926-5805.
\newblock \doi{https://doi.org/10.1016/j.autcon.2020.103078}.
\newblock URL \url{https://www.sciencedirect.com/science/article/pii/S0926580519305813}.

\bibitem[Liu et~al.(2023)Liu, Jiang, Zhang, Liu, Zhang, Biswas, and Stone]{llm+p}
Bo~Liu, Yuqian Jiang, Xiaohan Zhang, Qiang Liu, Shiqi Zhang, Joydeep Biswas, and Peter Stone.
\newblock Llm+p: Empowering large language models with optimal planning proficiency, 2023.

\bibitem[Mirzaee \& Kordjamshidi(2022)Mirzaee and Kordjamshidi]{SQA}
Roshanak Mirzaee and Parisa Kordjamshidi.
\newblock Transfer learning with synthetic corpora for spatial role labeling and reasoning.
\newblock In \emph{Proceedings of the 2022 Conference on Empirical Methods in Natural Language Processing}, pp.\  6148--6165, Abu Dhabi, United Arab Emirates, December 2022. Association for Computational Linguistics.
\newblock \doi{10.18653/v1/2022.emnlp-main.413}.
\newblock URL \url{https://aclanthology.org/2022.emnlp-main.413}.

\bibitem[Mirzaee et~al.(2021)Mirzaee, Rajaby~Faghihi, Ning, and Kordjamshidi]{mirzaee-etal-2021-spartqa}
Roshanak Mirzaee, Hossein Rajaby~Faghihi, Qiang Ning, and Parisa Kordjamshidi.
\newblock {SPARTQA}: A textual question answering benchmark for spatial reasoning.
\newblock In \emph{Proceedings of the 2021 Conference of the North American Chapter of the Association for Computational Linguistics: Human Language Technologies}, pp.\  4582--4598, Online, June 2021. Association for Computational Linguistics.
\newblock \doi{10.18653/v1/2021.naacl-main.364}.
\newblock URL \url{https://aclanthology.org/2021.naacl-main.364}.

\bibitem[Nguyen et~al.(2019)Nguyen, Dey, Brockett, and Dolan]{8954259}
Khanh Nguyen, Debadeepta Dey, Chris Brockett, and Bill Dolan.
\newblock Vision-based navigation with language-based assistance via imitation learning with indirect intervention.
\newblock In \emph{2019 IEEE/CVF Conference on Computer Vision and Pattern Recognition (CVPR)}, pp.\  12519--12529, 2019.
\newblock \doi{10.1109/CVPR.2019.01281}.

\bibitem[{Nye} et~al.(2021){Nye}, {Andreassen}, {Gur-Ari}, {Michalewski}, {Austin}, {Bieber}, {Dohan}, {Lewkowycz}, {Bosma}, {Luan}, {Sutton}, and {Odena}]{2021arXiv211200114N}
Maxwell {Nye}, Anders~Johan {Andreassen}, Guy {Gur-Ari}, Henryk {Michalewski}, Jacob {Austin}, David {Bieber}, David {Dohan}, Aitor {Lewkowycz}, Maarten {Bosma}, David {Luan}, Charles {Sutton}, and Augustus {Odena}.
\newblock {Show Your Work: Scratchpads for Intermediate Computation with Language Models}.
\newblock \emph{arXiv e-prints}, art. arXiv:2112.00114, November 2021.
\newblock \doi{10.48550/arXiv.2112.00114}.

\bibitem[Panov et~al.(2018)Panov, Yakovlev, and Suvorov]{PANOV2018347}
Aleksandr~I. Panov, Konstantin~S. Yakovlev, and Roman Suvorov.
\newblock Grid path planning with deep reinforcement learning: Preliminary results.
\newblock \emph{Procedia Computer Science}, 123:\penalty0 347--353, 2018.
\newblock ISSN 1877-0509.
\newblock \doi{https://doi.org/10.1016/j.procs.2018.01.054}.
\newblock URL \url{https://www.sciencedirect.com/science/article/pii/S1877050918300553}.
\newblock 8th Annual International Conference on Biologically Inspired Cognitive Architectures, BICA 2017 (Eighth Annual Meeting of the BICA Society), held August 1-6, 2017 in Moscow, Russia.

\bibitem[{Peng} et~al.(2023){Peng}, {Ding}, {Zhong}, {Shen}, {Liu}, {Zhang}, {Ouyang}, and {Tao}]{2023arXiv230313780P}
Keqin {Peng}, Liang {Ding}, Qihuang {Zhong}, Li~{Shen}, Xuebo {Liu}, Min {Zhang}, Yuanxin {Ouyang}, and Dacheng {Tao}.
\newblock {Towards Making the Most of ChatGPT for Machine Translation}.
\newblock \emph{arXiv e-prints}, art. arXiv:2303.13780, March 2023.

\bibitem[Rojowiec et~al.(2020)Rojowiec, G{\"o}tze, Sadler, Voigt, Zarrie{\ss}, and Schlangen]{rojowiec-etal-2020-generating}
Robin Rojowiec, Jana G{\"o}tze, Philipp Sadler, Henrik Voigt, Sina Zarrie{\ss}, and David Schlangen.
\newblock From {``}before{''} to {``}after{''}: Generating natural language instructions from image pairs in a simple visual domain.
\newblock In \emph{Proceedings of the 13th International Conference on Natural Language Generation}, pp.\  316--326, Dublin, Ireland, December 2020. Association for Computational Linguistics.
\newblock URL \url{https://aclanthology.org/2020.inlg-1.38}.

\bibitem[{Shi} et~al.(2023){Shi}, {Chen}, {Misra}, {Scales}, {Dohan}, {Chi}, {Sch{\"a}rli}, and {Zhou}]{2023arXiv230200093S}
Freda {Shi}, Xinyun {Chen}, Kanishka {Misra}, Nathan {Scales}, David {Dohan}, Ed~{Chi}, Nathanael {Sch{\"a}rli}, and Denny {Zhou}.
\newblock {Large Language Models Can Be Easily Distracted by Irrelevant Context}.
\newblock \emph{arXiv e-prints}, art. arXiv:2302.00093, January 2023.
\newblock \doi{10.48550/arXiv.2302.00093}.

\bibitem[Shi et~al.(2023)Shi, Suzgun, Freitag, Wang, Srivats, Vosoughi, Chung, Tay, Ruder, Zhou, Das, and Wei]{shi2023language}
Freda Shi, Mirac Suzgun, Markus Freitag, Xuezhi Wang, Suraj Srivats, Soroush Vosoughi, Hyung~Won Chung, Yi~Tay, Sebastian Ruder, Denny Zhou, Dipanjan Das, and Jason Wei.
\newblock Language models are multilingual chain-of-thought reasoners.
\newblock In \emph{The Eleventh International Conference on Learning Representations}, 2023.

\bibitem[{Srivastava} et~al.(2022){Srivastava}, {Rastogi}, {Rao}, {Shoeb}, {Abid}, {Fisch}, {Brown}, {Santoro}, {Gupta}, {Garriga-Alonso}, {Kluska}, {Lewkowycz}, {Agarwal}, {Power}, {Ray}, {Warstadt}, {Kocurek}, {Safaya}, {Tazarv}, {Xiang}, {Parrish}, {Nie}, {Hussain}, {Askell}, {Dsouza}, {Slone}, {Rahane}, {Iyer}, {Andreassen}, {Madotto}, {Santilli}, {Stuhlm{\"u}ller}, {Dai}, {La}, {Lampinen}, {Zou}, {Jiang}, {Chen}, {Vuong}, {Gupta}, {Gottardi}, {Norelli}, {Venkatesh}, {Gholamidavoodi}, {Tabassum}, {Menezes}, {Kirubarajan}, {Mullokandov}, {Sabharwal}, {Herrick}, {Efrat}, {Erdem}, {Karaka{\c{s}}}, {Roberts}, {Loe}, {Zoph}, {Bojanowski}, {{\"O}zyurt}, {Hedayatnia}, {Neyshabur}, {Inden}, {Stein}, {Ekmekci}, {Yuchen Lin}, {Howald}, {Diao}, {Dour}, {Stinson}, {Argueta}, {Ferri Ram{\'\i}rez}, {Singh}, {Rathkopf}, {Meng}, {Baral}, {Wu}, {Callison-Burch}, {Waites}, {Voigt}, {Manning}, {Potts}, {Ramirez}, {Rivera}, {Siro}, {Raffel}, {Ashcraft}, {Garbacea}, {Sileo}, {Garrette}, {Hendrycks}, {Kilman}, {Roth},
  {Freeman}, {Khashabi}, {Levy}, {Mosegu{\'\i} Gonz{\'a}lez}, {Perszyk}, {Hernandez}, {Chen}, {Ippolito}, {Gilboa}, {Dohan}, {Drakard}, {Jurgens}, {Datta}, {Ganguli}, {Emelin}, {Kleyko}, {Yuret}, {Chen}, {Tam}, {Hupkes}, {Misra}, {Buzan}, {Coelho Mollo}, {Yang}, {Lee}, {Shutova}, {Dogus Cubuk}, {Segal}, {Hagerman}, {Barnes}, {Donoway}, {Pavlick}, {Rodola}, {Lam}, {Chu}, {Tang}, {Erdem}, {Chang}, {Chi}, {Dyer}, {Jerzak}, {Kim}, {Engefu Manyasi}, {Zheltonozhskii}, {Xia}, {Siar}, {Mart{\'\i}nez-Plumed}, {Happ{\'e}}, {Chollet}, {Rong}, {Mishra}, {Indra Winata}, {de Melo}, {Kruszewski}, {Parascandolo}, {Mariani}, {Wang}, {Jaimovitch-L{\'o}pez}, {Betz}, {Gur-Ari}, {Galijasevic}, {Kim}, {Rashkin}, {Hajishirzi}, {Mehta}, {Bogar}, {Shevlin}, {Sch{\"u}tze}, {Yakura}, {Zhang}, {Wong}, {Ng}, {Noble}, {Jumelet}, {Geissinger}, {Kernion}, {Hilton}, {Lee}, {Fern{\'a}ndez Fisac}, {Simon}, {Koppel}, {Zheng}, {Zou}, {Koco{\'n}}, {Thompson}, {Kaplan}, {Radom}, {Sohl-Dickstein}, {Phang}, {Wei}, {Yosinski}, {Novikova}, {Bosscher},
  {Marsh}, {Kim}, {Taal}, {Engel}, {Alabi}, {Xu}, {Song}, {Tang}, {Waweru}, {Burden}, {Miller}, {Balis}, {Berant}, {Frohberg}, {Rozen}, {Hernandez-Orallo}, {Boudeman}, {Jones}, {Tenenbaum}, {Rule}, {Chua}, {Kanclerz}, {Livescu}, {Krauth}, {Gopalakrishnan}, {Ignatyeva}, {Markert}, {Dhole}, {Gimpel}, {Omondi}, {Mathewson}, {Chiafullo}, {Shkaruta}, {Shridhar}, {McDonell}, {Richardson}, {Reynolds}, {Gao}, {Zhang}, {Dugan}, {Qin}, {Contreras-Ochando}, {Morency}, {Moschella}, {Lam}, {Noble}, {Schmidt}, {He}, {Oliveros Col{\'o}n}, {Metz}, {Kerem {\c{S}}enel}, {Bosma}, {Sap}, {ter Hoeve}, {Farooqi}, {Faruqui}, {Mazeika}, {Baturan}, {Marelli}, {Maru}, {Ram{\'\i}rez Quintana}, {Tolkiehn}, {Giulianelli}, {Lewis}, {Potthast}, {Leavitt}, {Hagen}, {Schubert}, {Orduna Baitemirova}, {Arnaud}, {McElrath}, {Yee}, {Cohen}, {Gu}, {Ivanitskiy}, {Starritt}, {Strube}, {Sw{\k{e}}drowski}, {Bevilacqua}, {Yasunaga}, {Kale}, {Cain}, {Xu}, {Suzgun}, {Tiwari}, {Bansal}, {Aminnaseri}, {Geva}, {Gheini}, {Varma T}, {Peng}, {Chi}, {Lee},
  {Gur-Ari Krakover}, {Cameron}, {Roberts}, {Doiron}, {Nangia}, {Deckers}, {Muennighoff}, {Shirish Keskar}, {Iyer}, {Constant}, {Fiedel}, {Wen}, {Zhang}, {Agha}, {Elbaghdadi}, {Levy}, {Evans}, {Moreno Casares}, {Doshi}, {Fung}, {Liang}, {Vicol}, {Alipoormolabashi}, {Liao}, {Liang}, {Chang}, {Eckersley}, {Mon Htut}, {Hwang}, {Mi{\l}kowski}, {Patil}, {Pezeshkpour}, {Oli}, {Mei}, {Lyu}, {Chen}, {Banjade}, {Etta Rudolph}, {Gabriel}, {Habacker}, {Risco Delgado}, {Milli{\`e}re}, {Garg}, {Barnes}, {Saurous}, {Arakawa}, {Raymaekers}, {Frank}, {Sikand}, {Novak}, {Sitelew}, {LeBras}, {Liu}, {Jacobs}, {Zhang}, {Salakhutdinov}, {Chi}, {Lee}, {Stovall}, {Teehan}, {Yang}, {Singh}, {Mohammad}, {Anand}, {Dillavou}, {Shleifer}, {Wiseman}, {Gruetter}, {Bowman}, {Schoenholz}, {Han}, {Kwatra}, {Rous}, {Ghazarian}, {Ghosh}, {Casey}, {Bischoff}, {Gehrmann}, {Schuster}, {Sadeghi}, {Hamdan}, {Zhou}, {Srivastava}, {Shi}, {Singh}, {Asaadi}, {Gu}, {Pachchigar}, {Toshniwal}, {Upadhyay}, {Shyamolima}, {Debnath}, {Shakeri}, {Thormeyer},
  {Melzi}, {Reddy}, {Priscilla Makini}, {Lee}, {Torene}, {Hatwar}, {Dehaene}, {Divic}, {Ermon}, {Biderman}, {Lin}, {Prasad}, {Piantadosi}, {Shieber}, {Misherghi}, {Kiritchenko}, {Mishra}, {Linzen}, {Schuster}, {Li}, {Yu}, {Ali}, {Hashimoto}, {Wu}, {Desbordes}, {Rothschild}, {Phan}, {Wang}, {Nkinyili}, {Schick}, {Kornev}, {Telleen-Lawton}, {Tunduny}, {Gerstenberg}, {Chang}, {Neeraj}, {Khot}, {Shultz}, {Shaham}, {Misra}, {Demberg}, {Nyamai}, {Raunak}, {Ramasesh}, {Uday Prabhu}, {Padmakumar}, {Srikumar}, {Fedus}, {Saunders}, {Zhang}, {Vossen}, {Ren}, {Tong}, {Zhao}, {Wu}, {Shen}, {Yaghoobzadeh}, {Lakretz}, {Song}, {Bahri}, {Choi}, {Yang}, {Hao}, {Chen}, {Belinkov}, {Hou}, {Hou}, {Bai}, {Seid}, {Zhao}, {Wang}, {Wang}, {Wang}, and {Wu}]{2022arXiv220604615S}
Aarohi {Srivastava}, Abhinav {Rastogi}, Abhishek {Rao}, Abu Awal~Md {Shoeb}, Abubakar {Abid}, Adam {Fisch}, Adam~R. {Brown}, Adam {Santoro}, Aditya {Gupta}, Adri{\`a} {Garriga-Alonso}, Agnieszka {Kluska}, Aitor {Lewkowycz}, Akshat {Agarwal}, Alethea {Power}, Alex {Ray}, Alex {Warstadt}, Alexander~W. {Kocurek}, Ali {Safaya}, Ali {Tazarv}, Alice {Xiang}, Alicia {Parrish}, Allen {Nie}, Aman {Hussain}, Amanda {Askell}, Amanda {Dsouza}, Ambrose {Slone}, Ameet {Rahane}, Anantharaman~S. {Iyer}, Anders {Andreassen}, Andrea {Madotto}, Andrea {Santilli}, Andreas {Stuhlm{\"u}ller}, Andrew {Dai}, Andrew {La}, Andrew {Lampinen}, Andy {Zou}, Angela {Jiang}, Angelica {Chen}, Anh {Vuong}, Animesh {Gupta}, Anna {Gottardi}, Antonio {Norelli}, Anu {Venkatesh}, Arash {Gholamidavoodi}, Arfa {Tabassum}, Arul {Menezes}, Arun {Kirubarajan}, Asher {Mullokandov}, Ashish {Sabharwal}, Austin {Herrick}, Avia {Efrat}, Aykut {Erdem}, Ayla {Karaka{\c{s}}}, B.~Ryan {Roberts}, Bao~Sheng {Loe}, Barret {Zoph}, Bart{\l}omiej {Bojanowski},
  Batuhan {{\"O}zyurt}, Behnam {Hedayatnia}, Behnam {Neyshabur}, Benjamin {Inden}, Benno {Stein}, Berk {Ekmekci}, Bill {Yuchen Lin}, Blake {Howald}, Cameron {Diao}, Cameron {Dour}, Catherine {Stinson}, Cedrick {Argueta}, C{\'e}sar {Ferri Ram{\'\i}rez}, Chandan {Singh}, Charles {Rathkopf}, Chenlin {Meng}, Chitta {Baral}, Chiyu {Wu}, Chris {Callison-Burch}, Chris {Waites}, Christian {Voigt}, Christopher~D. {Manning}, Christopher {Potts}, Cindy {Ramirez}, Clara~E. {Rivera}, Clemencia {Siro}, Colin {Raffel}, Courtney {Ashcraft}, Cristina {Garbacea}, Damien {Sileo}, Dan {Garrette}, Dan {Hendrycks}, Dan {Kilman}, Dan {Roth}, Daniel {Freeman}, Daniel {Khashabi}, Daniel {Levy}, Daniel {Mosegu{\'\i} Gonz{\'a}lez}, Danielle {Perszyk}, Danny {Hernandez}, Danqi {Chen}, Daphne {Ippolito}, Dar {Gilboa}, David {Dohan}, David {Drakard}, David {Jurgens}, Debajyoti {Datta}, Deep {Ganguli}, Denis {Emelin}, Denis {Kleyko}, Deniz {Yuret}, Derek {Chen}, Derek {Tam}, Dieuwke {Hupkes}, Diganta {Misra}, Dilyar {Buzan}, Dimitri
  {Coelho Mollo}, Diyi {Yang}, Dong-Ho {Lee}, Ekaterina {Shutova}, Ekin {Dogus Cubuk}, Elad {Segal}, Eleanor {Hagerman}, Elizabeth {Barnes}, Elizabeth {Donoway}, Ellie {Pavlick}, Emanuele {Rodola}, Emma {Lam}, Eric {Chu}, Eric {Tang}, Erkut {Erdem}, Ernie {Chang}, Ethan~A. {Chi}, Ethan {Dyer}, Ethan {Jerzak}, Ethan {Kim}, Eunice {Engefu Manyasi}, Evgenii {Zheltonozhskii}, Fanyue {Xia}, Fatemeh {Siar}, Fernando {Mart{\'\i}nez-Plumed}, Francesca {Happ{\'e}}, Francois {Chollet}, Frieda {Rong}, Gaurav {Mishra}, Genta {Indra Winata}, Gerard {de Melo}, Germ{\'a}n {Kruszewski}, Giambattista {Parascandolo}, Giorgio {Mariani}, Gloria {Wang}, Gonzalo {Jaimovitch-L{\'o}pez}, Gregor {Betz}, Guy {Gur-Ari}, Hana {Galijasevic}, Hannah {Kim}, Hannah {Rashkin}, Hannaneh {Hajishirzi}, Harsh {Mehta}, Hayden {Bogar}, Henry {Shevlin}, Hinrich {Sch{\"u}tze}, Hiromu {Yakura}, Hongming {Zhang}, Hugh~Mee {Wong}, Ian {Ng}, Isaac {Noble}, Jaap {Jumelet}, Jack {Geissinger}, Jackson {Kernion}, Jacob {Hilton}, Jaehoon {Lee}, Jaime
  {Fern{\'a}ndez Fisac}, James~B. {Simon}, James {Koppel}, James {Zheng}, James {Zou}, Jan {Koco{\'n}}, Jana {Thompson}, Jared {Kaplan}, Jarema {Radom}, Jascha {Sohl-Dickstein}, Jason {Phang}, Jason {Wei}, Jason {Yosinski}, Jekaterina {Novikova}, Jelle {Bosscher}, Jennifer {Marsh}, Jeremy {Kim}, Jeroen {Taal}, Jesse {Engel}, Jesujoba {Alabi}, Jiacheng {Xu}, Jiaming {Song}, Jillian {Tang}, Joan {Waweru}, John {Burden}, John {Miller}, John~U. {Balis}, Jonathan {Berant}, J{\"o}rg {Frohberg}, Jos {Rozen}, Jose {Hernandez-Orallo}, Joseph {Boudeman}, Joseph {Jones}, Joshua~B. {Tenenbaum}, Joshua~S. {Rule}, Joyce {Chua}, Kamil {Kanclerz}, Karen {Livescu}, Karl {Krauth}, Karthik {Gopalakrishnan}, Katerina {Ignatyeva}, Katja {Markert}, Kaustubh~D. {Dhole}, Kevin {Gimpel}, Kevin {Omondi}, Kory {Mathewson}, Kristen {Chiafullo}, Ksenia {Shkaruta}, Kumar {Shridhar}, Kyle {McDonell}, Kyle {Richardson}, Laria {Reynolds}, Leo {Gao}, Li~{Zhang}, Liam {Dugan}, Lianhui {Qin}, Lidia {Contreras-Ochando}, Louis-Philippe {Morency},
  Luca {Moschella}, Lucas {Lam}, Lucy {Noble}, Ludwig {Schmidt}, Luheng {He}, Luis {Oliveros Col{\'o}n}, Luke {Metz}, L{\"u}tfi {Kerem {\c{S}}enel}, Maarten {Bosma}, Maarten {Sap}, Maartje {ter Hoeve}, Maheen {Farooqi}, Manaal {Faruqui}, Mantas {Mazeika}, Marco {Baturan}, Marco {Marelli}, Marco {Maru}, Maria~Jose {Ram{\'\i}rez Quintana}, Marie {Tolkiehn}, Mario {Giulianelli}, Martha {Lewis}, Martin {Potthast}, Matthew~L. {Leavitt}, Matthias {Hagen}, M{\'a}ty{\'a}s {Schubert}, Medina {Orduna Baitemirova}, Melody {Arnaud}, Melvin {McElrath}, Michael~A. {Yee}, Michael {Cohen}, Michael {Gu}, Michael {Ivanitskiy}, Michael {Starritt}, Michael {Strube}, Micha{\l} {Sw{\k{e}}drowski}, Michele {Bevilacqua}, Michihiro {Yasunaga}, Mihir {Kale}, Mike {Cain}, Mimee {Xu}, Mirac {Suzgun}, Mo~{Tiwari}, Mohit {Bansal}, Moin {Aminnaseri}, Mor {Geva}, Mozhdeh {Gheini}, Mukund {Varma T}, Nanyun {Peng}, Nathan {Chi}, Nayeon {Lee}, Neta {Gur-Ari Krakover}, Nicholas {Cameron}, Nicholas {Roberts}, Nick {Doiron}, Nikita {Nangia},
  Niklas {Deckers}, Niklas {Muennighoff}, Nitish {Shirish Keskar}, Niveditha~S. {Iyer}, Noah {Constant}, Noah {Fiedel}, Nuan {Wen}, Oliver {Zhang}, Omar {Agha}, Omar {Elbaghdadi}, Omer {Levy}, Owain {Evans}, Pablo~Antonio {Moreno Casares}, Parth {Doshi}, Pascale {Fung}, Paul~Pu {Liang}, Paul {Vicol}, Pegah {Alipoormolabashi}, Peiyuan {Liao}, Percy {Liang}, Peter {Chang}, Peter {Eckersley}, Phu {Mon Htut}, Pinyu {Hwang}, Piotr {Mi{\l}kowski}, Piyush {Patil}, Pouya {Pezeshkpour}, Priti {Oli}, Qiaozhu {Mei}, Qing {Lyu}, Qinlang {Chen}, Rabin {Banjade}, Rachel {Etta Rudolph}, Raefer {Gabriel}, Rahel {Habacker}, Ram{\'o}n {Risco Delgado}, Rapha{\"e}l {Milli{\`e}re}, Rhythm {Garg}, Richard {Barnes}, Rif~A. {Saurous}, Riku {Arakawa}, Robbe {Raymaekers}, Robert {Frank}, Rohan {Sikand}, Roman {Novak}, Roman {Sitelew}, Ronan {LeBras}, Rosanne {Liu}, Rowan {Jacobs}, Rui {Zhang}, Ruslan {Salakhutdinov}, Ryan {Chi}, Ryan {Lee}, Ryan {Stovall}, Ryan {Teehan}, Rylan {Yang}, Sahib {Singh}, Saif~M. {Mohammad}, Sajant {Anand},
  Sam {Dillavou}, Sam {Shleifer}, Sam {Wiseman}, Samuel {Gruetter}, Samuel~R. {Bowman}, Samuel~S. {Schoenholz}, Sanghyun {Han}, Sanjeev {Kwatra}, Sarah~A. {Rous}, Sarik {Ghazarian}, Sayan {Ghosh}, Sean {Casey}, Sebastian {Bischoff}, Sebastian {Gehrmann}, Sebastian {Schuster}, Sepideh {Sadeghi}, Shadi {Hamdan}, Sharon {Zhou}, Shashank {Srivastava}, Sherry {Shi}, Shikhar {Singh}, Shima {Asaadi}, Shixiang~Shane {Gu}, Shubh {Pachchigar}, Shubham {Toshniwal}, Shyam {Upadhyay}, {Shyamolima}, {Debnath}, Siamak {Shakeri}, Simon {Thormeyer}, Simone {Melzi}, Siva {Reddy}, Sneha {Priscilla Makini}, Soo-Hwan {Lee}, Spencer {Torene}, Sriharsha {Hatwar}, Stanislas {Dehaene}, Stefan {Divic}, Stefano {Ermon}, Stella {Biderman}, Stephanie {Lin}, Stephen {Prasad}, Steven~T. {Piantadosi}, Stuart~M. {Shieber}, Summer {Misherghi}, Svetlana {Kiritchenko}, Swaroop {Mishra}, Tal {Linzen}, Tal {Schuster}, Tao {Li}, Tao {Yu}, Tariq {Ali}, Tatsu {Hashimoto}, Te-Lin {Wu}, Th{\'e}o {Desbordes}, Theodore {Rothschild}, Thomas {Phan},
  Tianle {Wang}, Tiberius {Nkinyili}, Timo {Schick}, Timofei {Kornev}, Timothy {Telleen-Lawton}, Titus {Tunduny}, Tobias {Gerstenberg}, Trenton {Chang}, Trishala {Neeraj}, Tushar {Khot}, Tyler {Shultz}, Uri {Shaham}, Vedant {Misra}, Vera {Demberg}, Victoria {Nyamai}, Vikas {Raunak}, Vinay {Ramasesh}, Vinay {Uday Prabhu}, Vishakh {Padmakumar}, Vivek {Srikumar}, William {Fedus}, William {Saunders}, William {Zhang}, Wout {Vossen}, Xiang {Ren}, Xiaoyu {Tong}, Xinran {Zhao}, Xinyi {Wu}, Xudong {Shen}, Yadollah {Yaghoobzadeh}, Yair {Lakretz}, Yangqiu {Song}, Yasaman {Bahri}, Yejin {Choi}, Yichi {Yang}, Yiding {Hao}, Yifu {Chen}, Yonatan {Belinkov}, Yu~{Hou}, Yufang {Hou}, Yuntao {Bai}, Zachary {Seid}, Zhuoye {Zhao}, Zijian {Wang}, Zijie~J. {Wang}, Zirui {Wang}, and Ziyi {Wu}.
\newblock {Beyond the Imitation Game: Quantifying and extrapolating the capabilities of language models}.
\newblock \emph{arXiv e-prints}, art. arXiv:2206.04615, June 2022.
\newblock \doi{10.48550/arXiv.2206.04615}.

\bibitem[Suhr et~al.(2017)Suhr, Lewis, Yeh, and Artzi]{nlvr}
Alane Suhr, Mike Lewis, James Yeh, and Yoav Artzi.
\newblock A corpus of natural language for visual reasoning.
\newblock In \emph{Proceedings of the 55th Annual Meeting of the Association for Computational Linguistics (Volume 2: Short Papers)}, pp.\  217--223, Vancouver, Canada, July 2017. Association for Computational Linguistics.
\newblock \doi{10.18653/v1/P17-2034}.
\newblock URL \url{https://aclanthology.org/P17-2034}.

\bibitem[Sun et~al.(2024)Sun, Chen, Xu, Cheng, Ma, Yin, Wang, Han, Zhu, Yuan, et~al.]{sun2024survey}
Qiushi Sun, Zhirui Chen, Fangzhi Xu, Kanzhi Cheng, Chang Ma, Zhangyue Yin, Jianing Wang, Chengcheng Han, Renyu Zhu, Shuai Yuan, et~al.
\newblock A survey of neural code intelligence: Paradigms, advances and beyond.
\newblock \emph{arXiv preprint arXiv:2403.14734}, 2024.

\bibitem[Touvron et~al.(2023)Touvron, Martin, Stone, Albert, Almahairi, Babaei, Bashlykov, Batra, Bhargava, Bhosale, Bikel, Blecher, Ferrer, Chen, Cucurull, Esiobu, Fernandes, Fu, Fu, Fuller, Gao, Goswami, Goyal, Hartshorn, Hosseini, Hou, Inan, Kardas, Kerkez, Khabsa, Kloumann, Korenev, Koura, Lachaux, Lavril, Lee, Liskovich, Lu, Mao, Martinet, Mihaylov, Mishra, Molybog, Nie, Poulton, Reizenstein, Rungta, Saladi, Schelten, Silva, Smith, Subramanian, Tan, Tang, Taylor, Williams, Kuan, Xu, Yan, Zarov, Zhang, Fan, Kambadur, Narang, Rodriguez, Stojnic, Edunov, and Scialom]{touvron2023llama}
Hugo Touvron, Louis Martin, Kevin Stone, Peter Albert, Amjad Almahairi, Yasmine Babaei, Nikolay Bashlykov, Soumya Batra, Prajjwal Bhargava, Shruti Bhosale, Dan Bikel, Lukas Blecher, Cristian~Canton Ferrer, Moya Chen, Guillem Cucurull, David Esiobu, Jude Fernandes, Jeremy Fu, Wenyin Fu, Brian Fuller, Cynthia Gao, Vedanuj Goswami, Naman Goyal, Anthony Hartshorn, Saghar Hosseini, Rui Hou, Hakan Inan, Marcin Kardas, Viktor Kerkez, Madian Khabsa, Isabel Kloumann, Artem Korenev, Punit~Singh Koura, Marie-Anne Lachaux, Thibaut Lavril, Jenya Lee, Diana Liskovich, Yinghai Lu, Yuning Mao, Xavier Martinet, Todor Mihaylov, Pushkar Mishra, Igor Molybog, Yixin Nie, Andrew Poulton, Jeremy Reizenstein, Rashi Rungta, Kalyan Saladi, Alan Schelten, Ruan Silva, Eric~Michael Smith, Ranjan Subramanian, Xiaoqing~Ellen Tan, Binh Tang, Ross Taylor, Adina Williams, Jian~Xiang Kuan, Puxin Xu, Zheng Yan, Iliyan Zarov, Yuchen Zhang, Angela Fan, Melanie Kambadur, Sharan Narang, Aurelien Rodriguez, Robert Stojnic, Sergey Edunov, and Thomas
  Scialom.
\newblock Llama 2: Open foundation and fine-tuned chat models, 2023.

\bibitem[{Tsai} et~al.(2023){Tsai}, {Zhou}, {Liu}, {Li}, {Yu}, and {Mei}]{2023arXiv230402868T}
Chen~Feng {Tsai}, Xiaochen {Zhou}, Sierra~S. {Liu}, Jing {Li}, Mo~{Yu}, and Hongyuan {Mei}.
\newblock {Can Large Language Models Play Text Games Well? Current State-of-the-Art and Open Questions}.
\newblock \emph{arXiv e-prints}, art. arXiv:2304.02868, April 2023.
\newblock \doi{10.48550/arXiv.2304.02868}.

\bibitem[Wake et~al.(2023)Wake, Kanehira, Sasabuchi, Takamatsu, and Ikeuchi]{wake2023chatgpt}
Naoki Wake, Atsushi Kanehira, Kazuhiro Sasabuchi, Jun Takamatsu, and Katsushi Ikeuchi.
\newblock Chatgpt empowered long-step robot control in various environments: A case application, 2023.

\bibitem[Wei et~al.(2022)Wei, Wang, Schuurmans, Bosma, brian ichter, Xia, Chi, Le, and Zhou]{wei2022chain}
Jason Wei, Xuezhi Wang, Dale Schuurmans, Maarten Bosma, brian ichter, Fei Xia, Ed~H. Chi, Quoc~V Le, and Denny Zhou.
\newblock Chain of thought prompting elicits reasoning in large language models.
\newblock In Alice~H. Oh, Alekh Agarwal, Danielle Belgrave, and Kyunghyun Cho (eds.), \emph{Advances in Neural Information Processing Systems}, 2022.

\bibitem[{Wei} et~al.(2023){Wei}, {Hou}, {Lampinen}, {Chen}, {Huang}, {Tay}, {Chen}, {Lu}, {Zhou}, {Ma}, and {Le}]{2023arXiv230508298W}
Jerry {Wei}, Le~{Hou}, Andrew {Lampinen}, Xiangning {Chen}, Da~{Huang}, Yi~{Tay}, Xinyun {Chen}, Yifeng {Lu}, Denny {Zhou}, Tengyu {Ma}, and Quoc~V. {Le}.
\newblock {Symbol tuning improves in-context learning in language models}.
\newblock \emph{arXiv e-prints}, art. arXiv:2305.08298, May 2023.
\newblock \doi{10.48550/arXiv.2305.08298}.

\end{thebibliography}
\bibliographystyle{colm2024_conference}

\appendix
\section{Appendix}
\subsection{Broader Impact}
\textsc{CoS} is a prompting technique that is easy to use, which can effectively improve the performance of complex planning with LLMS. It also indicates that future training with LLMs can also be well benefited by employing \textsc{CoS} in the training procedure to further improve LLM's planning abilities. 
\subsection{Extended Settings}
\subsubsection{Number of Tokens}
We have mentioned that we used white spacing for calculating the number of tokens in the intermediate thinking steps. This was a typo and in fact, we accurately measures the number of tokens using the OpenAI Playground.\footnote{https://platform.openai.com/playground} The numbers we reported are correct and there is no need for modification.
\subsubsection{Designing the Intermediate Steps}
The intermediate steps we use in the demonstrations for CoT are created and modified from the zero-shot CoT from the LLMs by simply adding “Let’s think step by step” before the answer. We then manually correct the intermediate steps from the outputs of using zero-shot CoT for further improvements. We attempted our best efforts in tuning the baselines, and we report the best results we achieved.
\subsection{Few-shot Exemplars}
In the remaining of this section, we demonstrate the few-shot exemplars used in the experiments in our study. We demonstrate the exemplars for both \textsc{CoS} and CoT.

\begin{table}[h!]
\captionof{table}{\label{exme4}
Few-shot exemplars for full Chain-of-Symbol prompt for brick 1D.
}
\vspace{3mm}
\tiny
\setlength\tabcolsep{13pt}
\setlength\aboverulesep{0pt}\setlength\belowrulesep{0pt}
\setcellgapes{2pt}\makegapedcells
\begin{tabularx}{\linewidth}{X}
\hline
\\\textbf{Question}:
There is a set of bricks. For brick B, the color is yellow. The yellow brick C is on top of the brick B . The yellow brick A is on top of the brick C . Now we have to get a specific brick. The bricks must now be grabbed from top to bottom, and if the lower brick is to be grabbed, the upper brick must be removed first. How to get brick B
\\\textbf{Answer}: \\\hlc[beaublue]{A//C \\
C//B\\
In sum: A//C//B
}\\ So we get the result as A, C, B.

\\\textbf{Question}:
There is a set of bricks. The yellow brick A is on top of the brick C . The yellow brick B is on top of the brick A . For the brick C, the color is white. Now we have to get a specific brick. The bricks must now be grabbed from top to bottom, and if the lower brick is to be grabbed, the upper brick must be removed first. How to get brick A?
\\\textbf{Answer}: \\\hlc[beaublue]{B//A\\
A//C \\
In sum: B//A//C
} \\So we get the result as B, A.

\\\textbf{Question}:
There is a set of bricks. The blue brick A is on top of the brick E . For the brick C, the color is blue. The blue brick D is on top of the brick B . The white brick E is on top of the brick D . The blue brick B is on top of the brick C . Now we have to get a specific brick. The bricks must now be grabbed from top to bottom, and if the lower brick is to be grabbed, the upper brick must be removed first. How to get brick D?
\\\textbf{Answer}: \\\hlc[beaublue]{A//E \\
E//D\\
D//B \\
B//C\\
In sum: A//E//D//B//C
} \\So we get the result as A, E, D.

\\\textbf{Question}:
There is a set of bricks. The white brick F is on top of the brick C . The white brick C is on top of the brick B . The yellow brick E is on top of the brick A . For the brick B, the color is white. The white brick D is on top of the brick F . The blue brick A is on top of the brick D . Now we have to get a specific brick. The bricks must now be grabbed from top to bottom, and if the lower brick is to be grabbed, the upper brick must be removed first. How to get brick F?
\\\textbf{Answer}: \\\hlc[beaublue]{E//A\\
A//D \\
D//F\\
F//C \\
C//B\\
In sum: E//A//D//F//C//B
} \\So we get the result as E, A, D, F.

\\\textbf{Question}:
There is a set of bricks. The white brick F is on top of the brick D . The yellow brick B is on top of the brick A . The blue brick D is on top of the brick C . The white brick E is on top of the brick G . For the brick A, the color is blue. The blue brick C is on top of the brick E . The white brick G is on top of the brick B . Now we have to get a specific brick. The bricks must now be grabbed from top to bottom, and if the lower brick is to be grabbed, the upper brick must be removed first. How to get brick G?
\\\textbf{Answer}: \\\hlc[beaublue]{A//B//G//E//C//D//F\\
D//F\\
C//D\\
E//C\\
G//E\\
} \\So we get the result as F, D, C, E, G.

\\\\
\hline
\end{tabularx}
\end{table}

\begin{table}[h!]
\captionof{table}{\label{exme8}
Few-shot exemplars for full \textsc{CoT} prompt for brick 1D.}

\vspace{3mm}
\small
\setlength\tabcolsep{13pt}
\setlength\aboverulesep{0pt}\setlength\belowrulesep{0pt}
\setcellgapes{2pt}\makegapedcells
\begin{tabularx}{\linewidth}{X}
\hline
\\\textbf{Question}:
There is a set of bricks. The yellow brick B is on top of the brick D . For the brick D, the color is white. The yellow brick A is on top of the brick C . The yellow brick E is on top of the brick A . The blue brick C is on top of the brick B . Now we have to get a specific brick. The bricks must now be grabbed from top to bottom, and if the lower brick is to be grabbed, the upper brick must be removed first. How to get brick D?
\\\textbf{Answer}: \hlc[beaublue]{
The bricks from bottom to top is B, D, C, A, E
1. Remove brick E from the top of brick A.
2. Remove brick A from the top of brick C.
3. Remove brick C from the top of brick B.
4. Now brick B is the topmost yellow brick and can be grabbed, but we need to remove it to get to brick D.
5. Remove brick B from the top of brick D.
6. Now brick D is the topmost white brick and can be grabbed.
} So we get the result as E, A, C, B, D.

\\\textbf{Question}:
There is a set of bricks. For the brick A, the color is blue. The white brick B is on top of the brick C . The blue brick C is on top of the brick A . Now we have to get a specific brick. The bricks must now be grabbed from top to bottom, and if the lower brick is to be grabbed, the upper brick must be removed first. How to get brick B?
\\\textbf{Answer}: \hlc[beaublue]{
The bricks from bottom to top is A, C, B
1. Remove brick C from the top of brick A.
2. Now brick A is the topmost brick and can be grabbed, but we need to remove it to get to brick B.
3. Remove brick A from the top of brick C.
4. Now brick B is the topmost white brick and can be grabbed.
} So we get the result as C, A, B.

\\\textbf{Question}:
There is a set of bricks. The white brick B is on top of the brick A . The white brick A is on top of the brick G . The white brick G is on top of the brick D . The blue brick D is on top of the brick F . The white brick H is on top of the brick B . For the brick C, the color is yellow. The white brick E is on top of the brick C . The white brick F is on top of the brick E . Now we have to get a specific brick. The bricks must now be grabbed from top to bottom, and if the lower brick is to be grabbed, the upper brick must be removed first. How to get brick F?
\\\textbf{Answer}: \hlc[beaublue]{
The bricks from bottom to top is C, E, F, D, G, A, B, H
1. Remove brick H from the top of brick B.
2. Remove brick B from the top of brick A.
3. Remove brick A from the top of brick G.
4. Remove brick G from the top of brick D.
5. Remove brick D from the top of brick F.
6. Now brick F is the topmost white brick and can be grabbed.
} So we get the result as H, B, A, G, D, F.

\\\textbf{Question}:
There is a set of bricks. The white brick B is on top of the brick C . The yellow brick D is on top of the brick A . For the brick A, the color is yellow. The yellow brick E is on top of the brick B . The white brick C is on top of the brick D . Now we have to get a specific brick. The bricks must now be grabbed from top to bottom, and if the lower brick is to be grabbed, the upper brick must be removed first. How to get brick D?
\\\textbf{Answer}: \hlc[beaublue]{
The bricks from bottom to top is A, D, C, B, E
1. Remove brick E from the top of brick B.
2. Remove brick B from the top of brick C.
3. Now brick C is the topmost white brick and can be grabbed, but we need to remove it to get to brick D.
4. Remove brick C from the top of brick D.
5. Now brick D is the topmost yellow brick and can be grabbed.
} So we get the result as E, B, C, D.

\\\textbf{Question}:
There is a set of bricks. The yellow brick C is on top of the brick B . The white brick B is on top of the brick A . For the brick A, the color is white. Now we have to get a specific brick. The bricks must now be grabbed from top to bottom, and if the lower brick is to be grabbed, the upper brick must be removed first. How to get brick B?
\\\textbf{Answer}: \hlc[beaublue]{
The bricks from bottom to top is A, B, C
1. Now brick C is the topmost brick and can be grabbed, but we need to remove it to get to brick B.
2. Now brick B is the topmost white brick and can be grabbed.
} So we get the result as C, B.

\\\\
\hline
\end{tabularx}
\end{table}

\begin{table}[h!]
\captionof{table}{\label{exme14}
Few-shot exemplars for full \textsc{CoS} prompt for brick 2D.
}
\vspace{3mm}
\tiny
\setlength\tabcolsep{13pt}
\setlength\aboverulesep{0pt}\setlength\belowrulesep{0pt}
\setcellgapes{2pt}\makegapedcells
\begin{tabularx}{\linewidth}{X}
\hline
\\\textbf{Question}: There is a set of bricks. There is a brick A. The brick B is in front of the brick A. The brick C is in front of the brick B. The brick D is on top of the brick B . The brick E is on top of the brick A . The brick F is on top of the brick C . The brick G is on top of the brick D . The brick H is on top of the brick G . Now we have to get a specific brick. The bricks must now be grabbed from top to bottom, and if the lower brick is to be grabbed, the upper brick must be removed first. How to get brick A?

\\\textbf{Answer}: \\\hlc[beaublue]{We can get\\ A//E, \\C//F,\\ B//D//G//H,}\\ So we get the result as E, A.

\\\textbf{Question}: There is a set of bricks. For the brick A. The yellow brick B is on top of the brick A . The blue brick C is on top of the brick B . The blue brick D is on top of the brick C . Now we have to get a specific brick. The bricks must now be grabbed from top to bottom, and if the lower brick is to be grabbed, the upper brick must be removed first. How to get brick B?

\\\textbf{Answer}: \\\hlc[beaublue]{We can get\\  A//B//C//D.}\\ So we get the result as D, C, B.

\\\textbf{Question}: There is a set of bricks. For the brick A, the color is blue. The yellow brick B is in front of the brick A. The yellow brick C is in front of the brick B. The white brick D is on top of the brick C . The white brick E is on top of the brick D . The yellow brick F is on top of the brick E . Now we have to get a specific brick. The bricks must now be grabbed from top to bottom, and if the lower brick is to be grabbed, the upper brick must be removed first. How to get brick C?

\\\textbf{Answer}:\\ \hlc[beaublue]{We can get \\ C->B->A,\\ C//D//E//F.}\\ So we get the result as F, E, D, C.

\\\textbf{Question}: There is a set of bricks. The brick P is on top of the brick R . The brick J is on top of the brick B . The brick D is on top of the brick P . The brick R is on top of the brick H . The brick K is in front of the brick M. The brick B is on top of the brick D . For the brick M, the color is blue. The brick C is on top of the brick J . The brick H is in front of the brick K. Now we have to get a specific brick. The bricks must now be grabbed from top to bottom, and if the lower brick is to be grabbed, the upper brick must be removed first. How to get brick M?

\\\textbf{Answer}: \\\hlc[beaublue]{We can get\\  H//R//P//D//B//J//C,\\ M-> K -> H.}\\ So we get the result as M directly.

\\\textbf{Question}: There is a set of bricks. The brick K is on top of the brick F . The brick M is in front of the brick F. The brick N is on top of the brick K . For the brick O, the color is blue. The brick G is on top of the brick A . The brick F is in front of the brick I. The brick I is in front of the brick O. The brick A is on top of the brick N . Now we have to get a specific brick. The bricks must now be grabbed from top to bottom, and if the lower brick is to be grabbed, the upper brick must be removed first. How to get brick O?

\\\textbf{Answer}:\\ \hlc[beaublue]{We can get \\ F//K//N//A//G, \\F->I->O,}\\ So we get the result as O directly.

\\\\
\hline
\end{tabularx}
\end{table}

\begin{table}[h!]
\captionof{table}{\label{exme15}
Few-shot exemplars for full \textsc{CoT} prompt for brick 2D.
}
\vspace{3mm}
\small
\setlength\tabcolsep{13pt}
\setlength\aboverulesep{0pt}\setlength\belowrulesep{0pt}
\setcellgapes{2pt}\makegapedcells
\begin{tabularx}{\linewidth}{X}
\hline
\\\textbf{Question}: There are a set of bricks. There is a brick A. The brick B is in front of the brick A. The brick C is in front of the brick B. The brick D is on top of the brick B . The brick E is on top of the brick A . The brick F is on top of the brick C . The brick G is on top of the brick D . The brick H is on top of the brick G . Now we have to get a specific brick. The bricks must now be grabbed from top to bottom, and if the lower brick is to be grabbed, the upper brick must be removed first. How to get brick A?

\\\textbf{Answer}: \hlc[beaublue]{Let's think step by step:\\
1. To get brick A, we find E is on top of A.\\
2. We find E is on the top.\\
3 We need to remove brick E first, as it is on top of brick A.\\
4. Brick A is now accessible and can be grabbed.\\
} So we get the result as E, A.

\\\textbf{Question}: There are a set of bricks. For the brick A. The yellow brick B is on top of the brick A . The blue brick C is on top of the brick B . The blue brick D is on top of the brick C . Now we have to get a specific brick. The bricks must now be grabbed from top to bottom, and if the lower brick is to be grabbed, the upper brick must be removed first. How to get brick B?

\\\textbf{Answer}: \hlc[beaublue]{Let's think step by step:
1. To get brick B, we find C is on top of B
2. We find D is on top of C
3. We find D is on the top
4. We need to remove brick D, as it is on top of brick C.
5. We need to remove brick C, as it is on top of brick B.
6. Brick B is now accessible and can be grabbed.
} So we get the result as D, C, B.

\\\textbf{Question}: There are a set of bricks. The brick P is on top of the brick R . The brick J is on top of the brick B . The brick D is on top of the brick P . The brick R is on top of the brick H . The brick K is in front of the brick M. The brick B is on top of the brick D . For the brick M, the color is blue. The brick C is on top of the brick J . The brick H is in front of the brick K. Now we have to get a specific brick. The bricks must now be grabbed from top to bottom, and if the lower brick is to be grabbed, the upper brick must be removed first. How to get brick M?

\\\textbf{Answer}: \hlc[beaublue]{Let's think step by step:
1. To get brick M, we find there is no brick on top of brick M,\\
} So we get the result as M directly.

\\\textbf{Question}: There are a set of bricks. The brick K is on top of the brick F . The brick M is in front of the brick F. The brick N is on top of the brick K . For the brick O, the color is blue. The brick G is on top of the brick A . The brick F is in front of the brick I. The brick I is in front of the brick O. The brick A is on top of the brick N . Now we have to get a specific brick. The bricks must now be grabbed from top to bottom, and if the lower brick is to be grabbed, the upper brick must be removed first. How to get brick O?

\\\textbf{Answer}: \hlc[beaublue]{Let's think step by step:
1. To get brick O, we find there is no brick on top of brick O,\\
} So we get the result as O directly.

\\\textbf{Question}: There are a set of bricks. For the brick A, the color is blue. The yellow brick B is in front of the brick A. The yellow brick C is in front of the brick B. The white brick D is on top of the brick C . The white brick E is on top of the brick D . The yellow brick F is on top of the brick E . Now we have to get a specific brick. The bricks must now be grabbed from top to bottom, and if the lower brick is to be grabbed, the upper brick must be removed first. How to get brick C?

\\\textbf{Answer}: \hlc[beaublue]{Let's think step by step:
1. To get brick C, we find C is in front of B.\\
2. We find D is on top of C.\\
3. We find E is on top of D.\\
4. We find F is on top of E.\\
5. We need to remove brick F, as it is on top of brick E.\\
6. We need to remove brick E, as it is on top of brick D.\\
7. We need to remove brick D, as it is on top of brick C.\\
8. Brick C is now accessible and can be grabbed.\\
} So we get the result as F, E, D, C.

\\\\
\hline
\end{tabularx}
\end{table}

\begin{table}[h!]
\captionof{table}{\label{exme16}
Few-shot exemplars for full \textsc{CoS} prompt for Natural Language Navigation
}
\vspace{3mm}
\small
\setlength\tabcolsep{13pt}
\setlength\aboverulesep{0pt}\setlength\belowrulesep{0pt}
\setcellgapes{2pt}\makegapedcells
\begin{tabularx}{\linewidth}{X}
\hline
\\\textbf{Question}: There is a set of roads and a set of landmarks. The start point is bank A. There is a road which is 200 meters long from bank A to bank C. There is a road which is 100 meters long from bank C to house H. There is a road which is 100 meters long from house H to cinema F. There is a road which is 200 meters long from cinema F to store B. There is a road which is 100 meters long from store B to store G. There is a road which is 200 meters long from bank C to house D. There is a road which is 200 meters long from house D to garden J. There is a road which is 100 meters long from bank A to cinema I. There is a road which is 100 meters long from cinema I to house E. 
From the start point, how to reach the nearest store?\\\textbf{Answer}: \hlc[beaublue]{
There are two stores on the map, store B and store G.\\
- bank A / bank C / house H / cinema F / store B (200 + 100 + 100 + 200 = 600)\\
- bank A / bank C / house H / cinema F / store B / store G (200 + 100 + 100 + 200 + 100 = 700)\\
So we get the result as bank A / bank C / house H / cinema F / store B.\\
} So the answer is ACHFB

\\\textbf{Question}: There is a set of roads and a set of landmarks. The start point is bank H. There is a road which is 100 meters long from bank H to store E. There is a road which is 200 meters long from store E to bank C. There is a road which is 100 meters long from bank C to house A. There is a road which is 100 meters long from house A to house F. There is a road which is 200 meters long from bank C to garden I. There is a road which is 200 meters long from store E to cinema G. There is a road which is 200 meters long from cinema G to school J. There is a road which is 200 meters long from bank H to school D. There is a road which is 100 meters long from school D to store B. 
From the start point, how to reach the nearest school?\\\textbf{Answer}: \hlc[beaublue]{
There are two schools on the map, school J and school D.
- bank H / store E / bank C / garden I / cinema G / school J (200 + 200 + 200 = 600)\\
- bank H / school D (200) \\
We get the result as bank H / school D.\\
} So the answer is HD

\\\textbf{Question}: There is a set of roads and a set of landmarks. The start point is garden B. There is a road which is 200 meters long from garden B to cinema D. There is a road which is 200 meters long from cinema D to school J. There is a road which is 100 meters long from garden B to school I. There is a road which is 100 meters long from school I to house C. There is a road which is 100 meters long from house C to bank F. There is a road which is 100 meters long from house C to cinema G. There is a road which is 200 meters long from cinema G to store H. There is a road which is 100 meters long from store H to cinema E. There is a road which is 200 meters long from cinema E to bank A. 
From the start point, how to reach the nearest school?\\\textbf{Answer}: \hlc[beaublue]{
There are two schools on the map, school J and school I.\\
- garden B / cinema D / school J (200+200 = 400)\\
- garden B / school I (100)\\
We get the result as garden B / school I (100).\\
} So the answer is BI

\\\\
\hline
\end{tabularx}
\end{table}

\begin{table}[h!]
\captionof{table}{\label{exme19}
Few-shot exemplars for full \textsc{CoT} prompt for Natural Language Navigation
}
\vspace{3mm}
\small
\setlength\tabcolsep{13pt}
\setlength\aboverulesep{0pt}\setlength\belowrulesep{0pt}
\setcellgapes{2pt}\makegapedcells
\begin{tabularx}{\linewidth}{X}
\hline
\\\textbf{Question}: There is a set of roads and a set of landmarks. The start point is bank A. There is a road which is 200 meters long from bank A to bank C. There is a road which is 100 meters long from bank C to house H. There is a road which is 100 meters long from house H to cinema F. There is a road which is 200 meters long from cinema F to store B. There is a road which is 100 meters long from store B to store G. There is a road which is 200 meters long from bank C to house D. There is a road which is 200 meters long from house D to garden J. There is a road which is 100 meters long from bank A to cinema I. There is a road which is 100 meters long from cinema I to house E. 
From the start point, how to reach the nearest store?\\\textbf{Answer}: \hlc[beaublue]{
Let's think step by step
1. Start at bank A.
2. There are two stores on the map, store B and store G.
3. The road from bank A to store B is from bank A to bank C to house H to cinema F to store B (200 meters + 100 meters + 100 meters + 200 meters = 600 meters)
4. The road from bank A to store B is shorter than the road from bank A to store G.
5. Therefore, the nearest store from the start point (bank A) is store B.
} So the answer is ACHFB.

\\\textbf{Question}: There is a set of roads and a set of landmarks. The start point is bank H. There is a road which is 100 meters long from bank H to store E. There is a road which is 200 meters long from store E to bank C. There is a road which is 100 meters long from bank C to house A. There is a road which is 100 meters long from house A to house F. There is a road which is 200 meters long from bank C to garden I. There is a road which is 200 meters long from store E to cinema G. There is a road which is 200 meters long from cinema G to school J. There is a road which is 200 meters long from bank H to school D. There is a road which is 100 meters long from school D to store B. 
From the start point, how to reach the nearest school?\\\textbf{Answer}: \hlc[beaublue]{
Let's think step by step
1. Start at bank H.
2. There are two schools on the map, school J and school D.
3. The road from bank H to school J is from bank H to store E to cinema G to school J (200 meters + 200 meters + 200 meters = 600 meters).
4. The road from bank H to school D is from bank H to school D (200 meters).
5. Therefore, the nearest school from the start point (bank H) is school D.
6. Follow the road from bank H to school D (200 meters).
} So the answer is HD

\\\textbf{Question}: There is a set of roads and a set of landmarks. The start point is garden B. There is a road which is 200 meters long from garden B to cinema D. There is a road which is 200 meters long from cinema D to school J. There is a road which is 100 meters long from garden B to school I. There is a road which is 100 meters long from school I to house C. There is a road which is 100 meters long from house C to bank F. There is a road which is 100 meters long from house C to cinema G. There is a road which is 200 meters long from cinema G to store H. There is a road which is 100 meters long from store H to cinema E. There is a road which is 200 meters long from cinema E to bank A. 
From the start point, how to reach the nearest school?\\\textbf{Answer}: \hlc[beaublue]{
Let's think step by step
1. Start at garden B.
2. There are two schools on the map, school J and school I.
3. The road from garden B to school J is from garden B to cinema D to school J (200 meters + 200 meters = 400 meters).
4. The road from garden B to school I is from garden B to school I (100 meters).
5. Therefore, the nearest school from the start point (garden B) is school I.
6. Follow the road from garden B to school I (100 meters).
} So the answer is BI

\\\\
\hline
\end{tabularx}
\end{table}

\begin{table}[h!]
\captionof{table}{\label{exme22}
Few-shot exemplars for full \textsc{CoS} prompt for NLVR-based Manipulation.
}
\vspace{3mm}
\scriptsize
\setlength\tabcolsep{13pt}
\setlength\aboverulesep{0pt}\setlength\belowrulesep{0pt}
\setcellgapes{2pt}\makegapedcells
\begin{tabularx}{\linewidth}{X}
\hline
\\\textbf{Story}: There are three boxes. In the left box, there are one middle square in yellow, one middle square in black, one small square in blue, one middle square in blue. In the middle box, there are one large square in blue, one middle square in blue, one middle square in black, one large triangle in black, one middle round in blue, one small square in yellow. In the right box, there are one large round in blue, one small triangle in yellow, one large triangle in blue.
Question:How to move all rounds to the middle box?
\\\textbf{Answer}: \hlc[beaublue]{
To move all rounds to the middle box, we need to identify all the round objects in the three boxes. They are:\\

- middle box: (middle, round, blue)\\
- right box: (large, round, blue)\\

} Then we can move the objects not in the middle box above to the middle box one by one:\\
- (large, round, blue)  middle - right 

\\\textbf{Story}: There are three boxes. In the left box, there are one small round in blue, one small round in blue, one large round in black, one large square in blue, one small round in yellow, one small round in yellow. In the middle box, there are one large square in blue, one middle triangle in yellow. In the right box, there are one large round in black, one large round in blue, one middle triangle in black, one middle triangle in black.
Question:How to move all squares to the right box?
\\\textbf{Answer}: \hlc[beaublue]{
To move all sqaure objects to the right box, we need to identify all the square objects in the three boxes. They are:\\

- left box: (large, square, blue)\\
- middle box: (large, square, blue)\\

} Then we can move the objects not in the right box above to the right box:\\

- (large, square, blue) left - right box\\
- (large, square, blue) middle - right box\\

\\\textbf{Story}: There are three boxes. In the left box, there are one large square in blue, one middle square in blue, one small round in blue, one middle triangle in blue, one middle round in yellow, one large square in yellow. In the middle box, there are one small round in yellow, one middle square in blue, one small triangle in black, one small square in black, one small triangle in yellow, one large round in black. In the right box, there are one small square in yellow, one small triangle in yellow, one middle triangle in black, one large round in yellow, one middle square in blue, one large square in yellow.
Question:How to move all black objects to the right box?
\\\textbf{Answer}: \hlc[beaublue]{
To move all black objects to the right box, we need to identify all the black objects in the three boxes. They are:

- middle box: \\
(small, triangle, black), \\(small, square, black), \\(large, round, black)

} Then we can move the objects not in the right box above to the right box:\\

- (small, triangle, black) middle - right\\
- (small, square, black) middle - right\\
- (large, round, black) middle - right\\

\\\textbf{Story}: There are three boxes. In the left box, there are one middle square in yellow, one middle square in black, one small square in blue, one middle square in blue. In the middle box, there are one large square in blue, one middle square in blue, one middle square in black, one large triangle in black, one middle round in blue, one small square in yellow. In the right box, there are one large round in blue, one small triangle in yellow, one large triangle in blue.
Question:How to move all rounds to the middle box?
\\\textbf{Answer}: \hlc[beaublue]{
To move all rounds to the middle box, we need to identify all the round objects in the three boxes. They are:

- middle box: (middle, round, blue)\\
- right box: (large, round, blue)\\

} Then we can move the objects not in the middle box above to the middle box:
\\
- (large, round, blue) right - middle \\

\\\textbf{Story}: There are three boxes. In the left box, there are one large round in black, one small square in blue, one small square in blue, one small triangle in yellow, one middle square in yellow. In the middle box, there are one large square in yellow, one middle triangle in blue, one large round in black. In the right box, there are one large square in blue, one large triangle in black, one middle triangle in black.
Question:How to move all black objects to the left box?
\\\textbf{Answer}: \hlc[beaublue]{
To move all black objects to the left box, we need to identify all the black objects in the three boxes. They are:\\

- left box: (large, round, black)\\
- middle box: (large, round, black), \\
- right box: (middle, triangle, black), (large, triangle, black)\\

} Then we can move the objects not in the left box above to the left box:\\

1. (large, round, black) middle - left\\
2. (middle, triangle, black) right - left\\
3. (large, triangle, black) right - left\\

\\\\
\hline
\end{tabularx}
\end{table}

\begin{table}[h!]
\captionof{table}{\label{exme25}
Few-shot exemplars for full \textsc{CoT} prompt for NLVR-based Manipulation.
}
\vspace{3mm}
\scriptsize
\setlength\tabcolsep{13pt}
\setlength\aboverulesep{0pt}\setlength\belowrulesep{0pt}
\setcellgapes{2pt}\makegapedcells
\begin{tabularx}{\linewidth}{X}
\hline
\\\textbf{Story}: There are three boxes. In the left box, there are one middle square in yellow, one middle square in black, one small square in blue, one middle square in blue. In the middle box, there are one large square in blue, one middle square in blue, one middle square in black, one large triangle in black, one middle round in blue, one small square in yellow. In the right box, there are one large round in blue, one small triangle in yellow, one large triangle in blue.
Question:How to move all rounds to the middle box?
\\\textbf{Answer}: \hlc[beaublue]{
To move all rounds to the middle box, we need to identify all the round objects in the three boxes. They are:

- One middle round in blue in the middle box
- One large round in blue in the right box

} Then we can move the boxes above not in the middle box to the middle box one by one:

1.  Move the large round in blue from the right box to the middle box.

Now all the rounds are in the middle box.

\\\textbf{Story}: There are three boxes. In the left box, there are one small round in blue, one small round in blue, one large round in black, one large square in blue, one small round in yellow, one small round in yellow. In the middle box, there are one large square in blue, one middle triangle in yellow. In the right box, there are one large round in black, one large round in blue, one middle triangle in black, one middle triangle in black.
Question:How to move all squares to the right box?
\\\textbf{Answer}: \hlc[beaublue]{
To move all sqaure objects to the right box, we need to identify all the square objects in the three boxes. They are:

- One large square in blue in the left box
- One large square in blue in the middle box

} Then we can move the box above with the square object to the right box:

1. Move the large square in blue from the left box to the right box.
2. Move the large square in blue from the middle box to the right box.

Now all the square objects are in the right box.

\\\textbf{Story}: There are three boxes. In the left box, there are one large square in blue, one middle square in blue, one small round in blue, one middle triangle in blue, one middle round in yellow, one large square in yellow. In the middle box, there are one small round in yellow, one middle square in blue, one small triangle in black, one small square in black, one small triangle in yellow, one large round in black. In the right box, there are one small square in yellow, one small triangle in yellow, one middle triangle in black, one large round in yellow, one middle square in blue, one large square in yellow.
Question:How to move all black objects to the right box?
\\\textbf{Answer}: \hlc[beaublue]{
To move all black objects to the right box, we need to identify all the black objects in the three boxes. They are:

- One small triangle in black in the middle box
- One middle triangle in black in the middle box
- One large round in black in the middle box

} Then we can move the boxes above not in the right box to the right box one by one:
1. Move the small triangle in black from the middle box to the right box.
2. Move the middle triangle in black from the middle box to the right box.
3. Move the large round in black from the middle box to the right box.

Now all the black objects are in the right box.

\\\textbf{Story}: There are three boxes. In the left box, there are one middle square in yellow, one middle square in black, one small square in blue, one middle square in blue. In the middle box, there are one large square in blue, one middle square in blue, one middle square in black, one large triangle in black, one middle round in blue, one small square in yellow. In the right box, there are one large round in blue, one small triangle in yellow, one large triangle in blue.
Question:How to move all rounds to the middle box?
\\\textbf{Answer}: \hlc[beaublue]{
To move all rounds to the middle box, we need to identify all the round objects in the three boxes. They are:

- One middle round in blue in the middle box
- One large round in blue in the right box

} Then we can move the boxes above not in the middle box to the middle box one by one:

1. Move the large round in blue from the right box to the middle box.

Now all the rounds are in the middle box.

\\\textbf{Story}: There are three boxes. In the left box, there are one large round in black, one small square in blue, one small square in blue, one small triangle in yellow, one middle square in yellow. In the middle box, there are one large square in yellow, one middle triangle in blue, one large round in black. In the right box, there are one large square in blue, one large triangle in black, one middle triangle in black.
Question:How to move all black objects to the left box?
\\\textbf{Answer}: \hlc[beaublue]{
To move all black objects to the left box, we need to identify all the black objects in the three boxes. They are:

- One large round in black in the left box
- One large round in black in the middle box
- One large triangle in black in the right box
- One middle triangle in black in the right box

} Then we can move the boxes above not in the left box to the left box one by one:

1. Move the large round in black from the middle box to the left box.
2. Move the large triangle in black from the right box to the left box.
3. Move the middle triangle in black from the right box to the left box.

Now all the black objects are in the left box.

\\\\
\hline
\end{tabularx}
\end{table}

\begin{table}[h!]
\captionof{table}{\label{exme28}
Few-shot exemplars for full \textsc{CoT} prompt for SPARTQA.
}
\vspace{3mm}
\tiny
\setlength\tabcolsep{13pt}
\setlength\aboverulesep{0pt}\setlength\belowrulesep{0pt}
\setcellgapes{2pt}\makegapedcells
\begin{tabularx}{\linewidth}{X}
\hline
\\Example 1:\\
Background: There are three blocks called A, B, and C. In block A, there is a small blue square far above a small blue circle. There is a small black triangle far to the right of both small blue objects. The black triange is near and above a medium blue square. The medium blue square is touching the right edge of A and is near and above a medium black circle. Near and to the right of the medium black circle is a medium yellow triangle. The yellow triangle is to the left of the small blue circle. In B, which is to the right of A, there is a large blue square near and to the left of a large yellow triangle. There is also a small yellow triangle far below the square. In C, which is to the right of B. there is a small yellow circle near and above a small yellow triangle. The small yellow triangle is touching the top of a medium black triangle. The black triangle is touching the right edge of C. There is a large blue circle far below the black triangle.\\
Question: Are all yellow objects in C above a blue thing?\\
Candidate Answers:["Yes", "No", "Do not Know"]\\
Answer:\\
\hlc[beaublue]{
Let's analyze the information provided step by step:
In block C:
    1. There a small yellow circle near and above a small yellow triangle.
    2. The small yellow triangle is touching the top of a medium black triangle.
    3. There is a large blue circle far below the black triangle.
Based on the provided information, the small yellow circle and the small yellow triangle in Block C are indeed above a blue thing, which is the large blue circle.}
So, the correct answer is: "Yes".
\\
Example 2:\\
Background: There are three blocks called A, B, and C. A is to the left of B and B is to the left of C. In A, there is a small black triangle near and ot the left of a large blue circle. The large blue circle is near and to the left of a small blue circle. The small blue square is far below a medium yellow square. In B, there is a small yellow circle. In C, there is a medium black triangle that is near and to the left of a small yellow circle. The small yellow circle is near and to the left of a blue triangle. There is a large yellow square touching the bottom of the small yellow circle. Near and below the large yellow square is a large black square.\\
Question: What is the relation between the large black thing and the small yellow thing?\\
Candidate Answers:["left", "right", "above", "below", "near to", "far from",  "touching", "Do Not Know"]\\
Answer:\\
\hlc[beaublue]{
Let's analyze the information provided step by step:
In block C:
    1. There is a medium black triangle that is near and to the left of a small yellow circle.
    2. The small yellow circle is near and to the left of a blue triangle.
    3. There is a large yellow square touching the bottom of the small yellow circle.
    4. Near and below the large yellow square is a large black square.
Now, let's determine the relation between the "large black thing" (large black square) and the "small yellow thing" (small yellow circle):
    1. The large black square is in block C, and it is described as being "near and below" the large yellow square.
    2. The small yellow circle is also in block C, and it is described as being "near and to the left of" the medium black triangle.
Based on the provided information, the large black square is below the large yellow square and the small yellow circle is near to the medium black triangle.}
So, the correct answer: "below" and "near to".
\\
Example 3:\\
Background: There are three blocks called A, B, and C. A is to the right of B and B is to the right of C. In A, there is a medium black circe that is far to the right of a medium blue square. The medium blue square is near and to the right of a small yellow circle. Far above the blue square is a medium yellow triangle. Near above the medium triangle is a small blue circle. In B, there is a large blue square that is touching the right edge of B. There is a medium black triangle that is touching the left side of the blue square. Near and above the medium black triangle is a large blue circle. Near and above the large blue circle is a large yellow triangle. In C, there is a small yellow triangle near and to the right of a medium black circle. The yellow triangle is touching the right edge of C. Far to the left of the medium black circle is a small blue square.\\
Question: What block does not have any objects touching the edge of it?\\
Candidate Answers: ["A", "B", "C"]\\
Answer:\\
\hlc[beaublue]{
Let's analyze the information provided step by step:
In block A:
    1. There is a medium black circle, which is far to the right of a medium blue square.
    2. The medium blue square is near and to the right of a small yellow circle.
    3. Far above the blue square is a medium yellow triangle.
    4. Near above the medium triangle is a small blue circle.
In block B:
    1. There is a large blue square that is touching the right edge of B.
    2. There is a medium black triangle that is touching the left side of the blue square.
    3. Near and above the medium black triangle is a large blue circle.
    4. Near and above the large blue circle is a large yellow triangle.
In block C:
    1. There is a small yellow triangle near and to the right of a medium black circle.
    2. The yellow triangle is touching the right edge of C.
    3. Far to the left of the medium black circle is a small blue square.
Based on the provided information, it's Block A, which does not have any objects touching its edge.}
So, the correct answer is: "A".
\\
Example 4:\\
Background: There are three blocks called A, B, and C. A is to the left of B and B is to the left of C. In A, there is a large black triangle far above a medium black triangle. The medium black triangle is near and above a large yellow circle. Near and to the right of the large circle is a small blue circle. In B, there is a medium blue circle near and to the left of a small black square. Far below the blue circle is a small yellow circle. In C, there is a small blue square.\\
Question: What object is near the large yellow thing, the medium black triangle or the small blue circle the medium black triangle  or  the smal blue circle?\\
Candidate Answers:["the medium black triangle", "the smal blue circle", "both of them", "none of them"]\\
Answer:\\
\hlc[beaublue]{
Let's analyze the information provided step by step:
In lock A:
    1.There is a large black triangle far above a medium black triangle. 
    2. The medium black triangle is near and above a large yellow circle.
    3. Near and to the right of the large circle is a small blue circle.
In block B:
    1. There is a medium blue circle near and to the left of a small black square. 
    2. Far below the blue circle is a small yellow circle.
In block C:
    1. There is a small blue square.
Now, let's analyze the position of the large yellow circle:
    1. The large yellow circle is in Block A and is near and above the medium black triangle.
    2. The small blue circle is also in Block A and is near and to the right of the large yellow circle.
Based on the provided information, both the medium black triangle and the small blue circle are near the large yellow circle in Block A.}
So, the correct answer is: "both of them".
\\
Example 5:\\
Background: There are three blue shapes in a block that we called A. Two medium blue squares are below a big blue square. First medium blue square is to the left of the other medium one which is touching the bottom edge of block A. There is another block below the block A we call it B. A big blue triangle is touching the top edge of this block which is above a small black triangle, a big blue circle and a medium blue triangle. The medium triangle is touching the bottom edge of the block. The circle is below and to the left of the small triangle. It is above the medium triangle.\\
Question: Is the small triangle below the medium triangle?\\
Candidate Answers:["Yes", "No", "Do not Know"]\\
Answer:\\\hlc[beaublue]{
Let's analyze the information provided step by step:
In block B:
    1. A big blue triangle is touching the top edge of this block which is above a small black triangle, a big blue circle and a medium blue triangle. 
    2. The medium triangle is touching the bottom edge of the block.
    3. The circle is below and to the left of the small triangle. It is above the medium triangle.
Based on the provided information, the small black triangle is above the medium blue triangle.}
So, the correct answer is: "No".
\\\\
\hline
\end{tabularx}
\end{table}

\begin{table}[h!]
\captionof{table}{\label{exme29}
Few-shot exemplars for full \textsc{CoS} prompt for SPARTQA.
}
\tiny
\setlength\tabcolsep{13pt}
\setlength\aboverulesep{0pt}\setlength\belowrulesep{0pt}
\setcellgapes{2pt}\makegapedcells
\begin{tabularx}{\linewidth}{X}
\hline
\\Example 1:\\
Background: There are three blocks called A, B, and C. In block A, there is a small blue square far above a small blue circle. There is a small black triangle far to the right of both small blue objects. The black triangle is near and above a medium blue square. The medium blue square is touching the right edge of A and is near and above a medium black circle. Near and to the right of the medium black circle is a medium yellow triangle. The yellow triangle is to the left of the small blue circle. In B, which is to the right of A, there is a large blue square near and to the left of a large yellow triangle. There is also a small yellow triangle far below the square. In C, which is to the right of B. There is a small yellow circle near and above a small yellow triangle. The small yellow triangle is touching the top of a medium black triangle. The black triangle is touching the right edge of C. There is a large blue circle far below the black triangle.\\
Question: Are all yellow objects in C above a blue thing?\\
Candidate Answers:["Yes", "NO", "Do not Know"]\\
Answer:\\
\hlc[beaublue]{
Let's analyze the information provided step by step to determine if all yellow objects in Block C are above a blue thing:
C: 
1. (small, yellow, circle) ~↑ (small, yellow, triangle)
2. (small, yellow, triangle) = top of (medium, black, triangle)
3. (large, blue, circle) $\infty$↓ (medium, black, triangle).
Based on the provided information, C: (small, yellow, circle) ↑ (large, blue, circle) and (small, yellow, triangle) ↑ (large, blue, circle).}
So, the correct answer is: "Yes".
\\
Example 2:\\
Background: There are three blocks called A, B, and C. A is to the left of B and B is to the left of C. In A, there is a small black triangle near and to the left of a large blue circle. The large blue circle is near and to the left of a small blue circle. The small blue square is far below a medium yellow square. In B, there is a small yellow circle. In C, there is a medium black triangle that is near and to the left of a small yellow circle. The small yellow circle is near and to the left of a blue triangle. There is a large yellow square touching the bottom of the small yellow circle. Near and below the large yellow square is a large black square.\\
Question: What is the relation between the large black thing and the small yellow thing?\\
Candidate Answers:["left", "right", "above", "below", "near to", "far from",  "touching", "Do Not Know"]\\
Answer:\\
\hlc[beaublue]{
Let's break down the relationships between the objects step by step:
A:
1. (small, black, triangle) ~< (large, blue, circle)
2. (large, blue, circle) ~< (small, blue, circle)
3. (small, blue, square) $\infty$↓ (medium, yellow, square)
B:
1. (small, yellow, circle)
C:
1. (medium, black, triangle) ~< (small, yellow, circle)
2. (small, yellow, circle) ~< (blue, triangle)
3. (large, yellow, square) =↓ (small, yellow, circle)
4. (large, black, square) ~↓ (large, yellow, square)
Now, let's determine the relation between the (large, black, square) and the (small, yellow, circle):
1. C: (large, black, square) ~↓ (large, yellow, square). Therefore, (large, black, square) ↓ (large, yellow, square)
2. C: (small, yellow, circle) ~< (medium, black, triangle). Therefore, (small, yellow, circle) ~ (medium, black, triangle)}
So, the correct answer: "below" and "near to".
\\
Example 3:\\
Background: There are three blocks called A, B, and C. A is to the right of B and B is to the right of C. In A, there is a medium black circle that is far to the right of a medium blue square. The medium blue square is near and to the right of a small yellow circle. Far above the blue square is a medium yellow triangle. Near the medium triangle is a small blue circle. In B, there is a large blue square that is touching the right edge of B. There is a medium black triangle that is touching the left side of the blue square. Near and above the medium black triangle is a large blue circle. Near and above the large blue circle is a large yellow triangle. In C, there is a small yellow triangle near and to the right of a medium black circle. The yellow triangle is touching the right edge of C. Far to the left of the medium black circle is a small blue square.\\
Question: What block does not have any objects touching the edge of it?\\
Candidate Answers:["A", "B", "C"]\\
Answer:\\\hlc[beaublue]{
Let's analyze each block step by step to determine which one does not have any objects touching its edge:
A:
1. (medium, black, circle) $\infty$> (medium, blue, square).
2. (medium, blue, square) ~> (small, yellow, circle).
3. (medium, yellow, triangle) $\infty$↑ (blue, square).
4. (small, blue, circle) ~↑ (medium, triangle).
B:
1. (large, blue, square) => edge of B.
2. (medium, black, triangle) =< (blue, square).
3. (large, blue, circle) ~↑ (medium, black, triangle).
4. (large, yellow, triangle) ~↑ (large, blue, circle).
C:
1. (small, yellow, triangle) ~> (medium, black, circle).
2. (yellow, triangle) => edge of C.
3. (small, blue, square) $\infty$< (medium, black, circle).
Based on the provided information,, let's determine which block does not have any objects touching its edge. It's Block B.}
So, the correct answer is: "A".
\\
Example 4:\\
Background: There are three blocks called A, B, and C. A is to the left of B and B is to the left of C. In A, there is a large black triangle far above a medium black triangle. The medium black triangle is near and above a large yellow circle. Near and to the right of the large circle is a small blue circle. In B, there is a medium blue circle near and to the left of a small black square. Far below the blue circle is a small yellow circle. In C, there is a small blue square.\\
Question: What object is near the large yellow thing, the medium black triangle the small blue circle the medium black triangle or the small blue circle?\\
Candidate Answers:["the medium black triangle", "the small blue circle", "both of them", "none of them"]\\
Answer:\\
\hlc[beaublue]{
Let's break down the information step by step to determine which object is near the large yellow circle, the medium black triangle, or the small blue circle:
A: 
1. (large, black, triangle) $\infty$↑ (medium, black, triangle). 
2. (medium, black, triangle) ~↑ (large, yellow, circle). 
3. (small, blue, circle) ~> (large, circle).
B: 
1. (medium, blue, circle) ~< (small, black, square). 
2. (small, yellow, circle) $\infty$↓ (blue, circle).
C:
1. there is a small blue square.
Now, let's analyze the position of the (large, yellow, circle):
1. A: (large, yellow, circle) ~↑ (medium, black, triangle).
2. A: (small, blue, circle) ~> (large, yellow, circle).
Based on the provided information, A: (medium, black, triangle) ~ (large, yellow, circle) and  (small, blue, circle) ~ (large, yellow, circle).}
So, the correct answer is: "both of them".\\
Example 5:\\
Background: There are three blue shapes in a block that we call A. Two medium blue squares are below a big blue square. The first medium blue square is to the left of the other medium one which is touching the bottom edge of block A. There is another block below block A we call it B. A big blue triangle is touching the top edge of this block which is above a small black triangle, a big blue circle and a medium blue triangle. The medium triangle is touching the bottom edge of the block. The circle is below and to the left of the small triangle. It is above the medium triangle.\\
Question: Is the small triangle below the medium triangle?\\
Candidate Answers:["Yes", "NO", "Do not Know"]\\
Answer:\\\hlc[beaublue]{
Let's analyze the arrangement of objects in block B to determine if the small triangle is below the medium triangle:
B:
1. (big, blue, triangle) = top edge of B
2. (big, blue, triangle) ↑ (small, black, triangle), (big, blue, circle) and (medium, blue, triangle)
2. (big, blue, circle) ↓< (small, black, triangle)
3. (big, blue, circle) ~↓ (big, blue, triangle)
4. (medium, blue, triangle) = bottom edge B
Based on the provided information, the small black triangle is above the medium blue triangle.} So, the correct answer is: "No."
\end{tabularx}
\end{table}

\begin{table}[h!]
\captionof{table}{\label{exme29}
Comparison of \textsc{CoS}(above) and \textsc{CoT}(below) prompts for BBH's Temporal Sequence task.
}
\tiny
\setlength\tabcolsep{13pt}
\setlength\aboverulesep{0pt}\setlength\belowrulesep{0pt}
\setcellgapes{2pt}\makegapedcells
\begin{tabularx}{\linewidth}{X}
\hline
\\Example 1:\\
Q: Today, Emily went to the museum. Between what times could they have gone?\\
We know that:\\
Emily woke up at 1pm.\\
Elizabeth saw Emily reading at the library from 2pm to 4pm.\\
Jessica saw Emily watching a movie at the theater from 4pm to 5pm.\\
Leslie saw Emily waiting at the airport from 5pm to 6pm.\\
William saw Emily buying clothes at the mall from 6pm to 7pm.\\
The museum was closed after 7pm.\\
Between what times could Emily have gone to the museum?\\
Options:\\
(A) 1pm to 2pm\\
(B) 6pm to 7pm\\
(C) 5pm to 6pm\\
(D) 2pm to 4pm\\
A: Let's think step by step.\\
\\\hlc[beaublue]{
1pm\textbf{/}woke-up.\\
1pm-2pm\textbf{//}free.\\
2pm-4pm\textbf{/}reading.\\
4pm-5pm\textbf{/}watching\\
5pm-6pm\textbf{/}waiting.\\
6pm-7pm\textbf{/}buying\\
7pm\textbf{/}closure.\\
=> 1pm-2pm//free, So the answer is (A).}\\
\\
\\
Q: Today, Emily went to the museum. Between what times could they have gone?\\
We know that:\\
Emily woke up at 1pm.\\
Elizabeth saw Emily reading at the library from 2pm to 4pm.\\
Jessica saw Emily watching a movie at the theater from 4pm to 5pm.\\
Leslie saw Emily waiting at the airport from 5pm to 6pm.\\
William saw Emily buying clothes at the mall from 6pm to 7pm.\\
The museum was closed after 7pm.\\
Between what times could Emily have gone to the museum?\\
Options:\\
(A) 1pm to 2pm\\
(B) 6pm to 7pm\\
(C) 5pm to 6pm\\
(D) 2pm to 4pm\\
A: Let's think step by step.\\
\\\hlc[beaublue]{Wake-up time: 1pm.\\
1pm-2pm: free.\\
2pm-4pm: reading at the library.\\
4pm-5pm: watching a movie at the theater.\\
5pm-6pm: waiting at the airport.\\
6pm-7pm: buying clothes at the mall.\\
The museum closure time: 7pm.\\
The only time when Emily could have gone to the museum was 1pm to 2pm. So the answer is (A).}\\

\end{tabularx}
\end{table}

\end{document}